\documentclass[10pt,twocolumn,letterpaper]{article}

\usepackage[pagenumbers]{cvpr}

\usepackage{titletoc}
\usepackage{xspace}
\usepackage{xcolor}
\usepackage{enumitem}
\usepackage{balance}
\definecolor{DarkGreen}{RGB}{15, 157, 88}
\definecolor{lightblue}{RGB}{220, 230, 250}
\definecolor{DarkBlue}{RGB}{15, 88, 157}

\usepackage{algorithm}
\usepackage{listings}
\usepackage{xcolor}
\usepackage{url}

\definecolor{codegreen}{rgb}{0,0.6,0}
\definecolor{codegray}{rgb}{0.5,0.5,0.5}
\definecolor{codepurple}{rgb}{0.58,0,0.82}
\definecolor{backcolour}{rgb}{0.95,0.95,0.92}

\lstdefinestyle{mystyle}{
    backgroundcolor=\color{backcolour},   
    commentstyle=\color{codegreen},
    keywordstyle=\color{magenta},
    numberstyle=\tiny\color{codegray},
    stringstyle=\color{codepurple},
    basicstyle=\ttfamily\scriptsize,
    breakatwhitespace=false,         
    breaklines=true,                 
    captionpos=b,                    
    keepspaces=true,                 
    numbers=left,                    
    numbersep=5pt,                  
    showspaces=false,                
    showstringspaces=false,
    showtabs=false,                  
    tabsize=2
}

\definecolor{cvprblue}{rgb}{0.21,0.49,0.74}
\usepackage[pagebackref,breaklinks,colorlinks,allcolors=cvprblue]{hyperref}
\usepackage{multirow}
\usepackage{siunitx}  
\usepackage{mdframed}
\usepackage{amsmath}
\usepackage{etoolbox}
\usepackage[table]{xcolor}

\makeatletter
\patchcmd{\paragraph}{\@startsection{paragraph}{4}{\z@}{3.25ex \@plus1ex \@minus.2ex}{-1em}}{\@startsection{paragraph}{4}{\z@}{0.5ex \@plus0.1ex \@minus.1ex}{-1em}}{}{}
\makeatother

\usepackage{xcolor}
\usepackage[most]{tcolorbox}
\usepackage{lipsum}

\AtBeginEnvironment{align}{\vspace{-2ex}}
\AtEndEnvironment{align}{\vspace{-1.5ex}}

\AtBeginEnvironment{equation}{\vspace{-2ex}}
\AtEndEnvironment{equation}{\vspace{-1.5ex}}

\definecolor{lightblue}{RGB}{230, 240, 255}
\definecolor{blushpink}{RGB}{255, 230, 230}
\definecolor{highlightcolor}{rgb}{0.8, 0.9, 1.0}

\newenvironment{qabox}[1]{%
 \vspace{4pt}
  \noindent\colorbox{lightblue}{%
    \parbox{\dimexpr\linewidth-2\fboxsep\relax}{%
      \textbf{Q. #1}
    }%
  }\par\vspace{1.5pt}
  \begin{tcolorbox}[
    colback=red!5, 
    colframe=black, 
    boxrule=0.5pt, 
    arc=2mm,
    top=2pt, bottom=2pt, left=2pt, right=2pt,
    boxsep=1pt,
    before skip=0pt,
    after skip=1ex
  ]
}{%
  \end{tcolorbox}%
}

\def\paperID{****}
\def\confName{CVPR}
\def\confYear{2026}

\title{Unique Lives, Shared World: Learning from Single-Life Videos}

\author{
Tengda Han\textsuperscript{$\star$}, Sayna Ebrahimi\textsuperscript{$\star$}, Dilara Gokay, Li Yang Ku, Maks Ovsjanikov, Iva Babukova, \\
Daniel Zoran, Viorica P{\u{a}}tr{\u{a}}ucean, Jo{\~a}o Carreira, Andrew Zisserman, Dima Damen\\
{\small Google DeepMind} \\
{\footnotesize\url{https://sites.google.com/view/learn-from-single-life}\vspace{-3mm}}
}

\newcommand{\cl}{Anonymous Lives}
\newcommand{\app}{Appendix}

\begin{document}
\maketitle
{
    \renewcommand{\thefootnote}{$\star$}
    \footnotetext{Equal contribution. First author order determined by coin flip.}
}
\begin{abstract}
We introduce the ``single-life” learning paradigm, where we train a distinct vision model exclusively on egocentric videos captured by one individual. We leverage the multiple viewpoints naturally captured within a single life to learn a visual encoder in a self-supervised manner. Our experiments demonstrate three key findings. First, models trained independently on different lives develop a highly aligned geometric understanding. We demonstrate this by training visual encoders on distinct datasets each capturing a different life, both indoors and outdoors, as well as introducing a novel cross-attention-based metric to quantify the functional alignment of the internal representations developed by different models. Second, we show that single-life models learn generalizable geometric representations that effectively transfer to downstream tasks, such as depth estimation, in unseen environments. Third, we demonstrate that training on up to 30 hours from one week of the same person's life leads to comparable performance to training on 30 hours of diverse web data, highlighting the strength of single-life representation learning.  Overall, our results establish that the shared structure of the world, both leads to consistency in models trained on individual lives, and provides a powerful signal for visual representation learning.
\end{abstract}

\vspace{-5mm}
\section{Introduction}
\label{sec:intro}
The goal of visual representation learning is to build models that can perceive 
the visual world in a generalizable way. Current representation learning approaches are built on the principles of data scale and diversity, and demonstrate remarkable generalization capabilities across a wide range of downstream tasks~\cite{oquab2023dinov2,gemini25,gpt4,xu2025qwen3,chen2024internvl}.  

This paradigm, however, differs significantly from that used in human learning. Living beings learn from a highly redundant stream of visual data derived \emph{only} from their own experiences, not from a randomized collection of images or clips gathered from unrelated sources.
Critically, we note that while individual visual experiences are distinct, they all arise from and are conditioned by the same underlying physical world. Structural properties such as 3D Euclidean geometry and object permanence are universals that leave a consistent imprint on all visual data. 

\begin{figure}[t]
    \centering
    \includegraphics[width=1\linewidth]{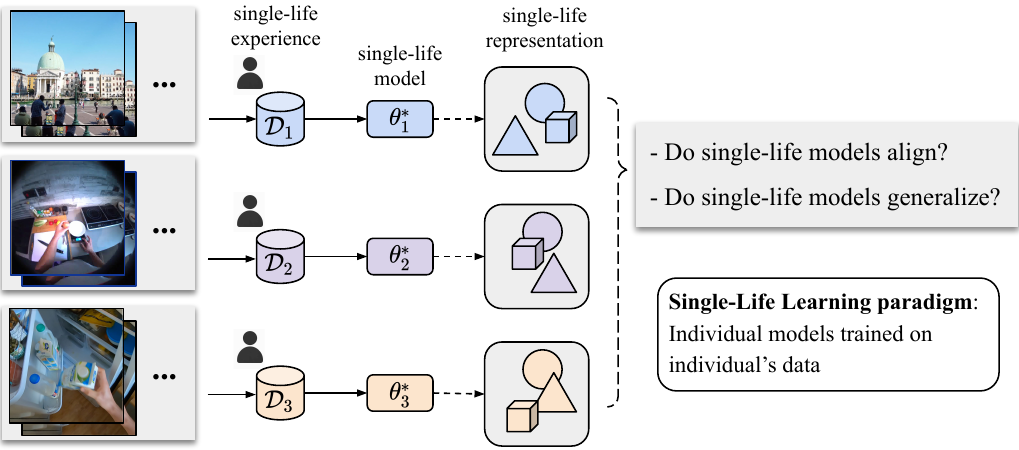}
    \caption{\textbf{Single-Life Learning Paradigm}. We train \emph{a distinct model} on egocentric video from one person's experience, and study the alignment and generalization of these models.
    }
    \label{fig:paradigm}
\end{figure}

In this paper, we introduce the ``single-life'' learning paradigm, where only visual data from one person's experience is used in learning.  We  define a `life' as the set of egocentric video data captured by a unique  participant as depicted in Fig.~\ref{fig:paradigm}. 
We use three datasets of increasing duration captured by individual participants (1 hour in~\cite{venkataramanan2024dora}, up to 7 hours in~\cite{perrett2025hdepic} and up to 38 hours recorded over one week in~\cite{adl}), to train 20 different `single life' models. 
While related to prior work on learning from video streams~\cite{carreira2024bl}, we are not concerned with the online, continuous nature of learning, but rather with the ability to learn robust visual representations using a single life as a self-contained source of data. We focus on geometric or structural downstream tasks, and thus adopt a proven self-supervised pipeline for learning geometry: cross-view completion \citep{weinzaepfel2022croco,gupta2023siamesemae}. The core principle is that observing a scene from multiple viewpoints provides a powerful signal for learning 3D structure. Within a single life, such diverse viewpoints are naturally captured as the individual moves through and re-visits their environment. This allows us to replicate the necessary conditions for geometric learning without aggregating data from multiple individuals.

Our main observation is that although each life is unique, visual experiences are not independent and all inherit the structural properties of the physical world. This notion is closely related to the Platonic Representation Hypothesis~\cite{huh2024platonic}, which posits that large models trained on web-scale data converge towards a single representation of reality. Adopting it to our setting, we formulate the \emph{Shared World Hypothesis}: because all lives are grounded in the same physical reality, geometric representations learned from these individual experiences should converge to a functionally similar structure.  Additionally, our work is inspired by infants' ability to develop geometric perception and 3D spatial awareness very early in life, significantly prior to semantic understanding \cite{soska2008development,choi2006influence}. We therefore test whether single-life videos can be used to train consistent and geometrically aware vision models without language supervision.

Our experiments reveal two remarkable properties. First, and most fundamentally, we show that models trained independently on individual lives arrive at a highly consistent and aligned geometric understanding, providing direct evidence for \emph{Shared World Hypothesis}. We demonstrate this by comparing properties of  various `single-life' datasets and then  introducing a new metric inspired by \cite{huh2024platonic} and \cite{an2025cross} to measure the alignment of learned representations using attention maps. Second, we show that these single-life models learn generalizable geometric representations that successfully transfer to unseen domains and downstream tasks.
In summary, our contributions are:
\begin{enumerate}[leftmargin=1.5em]
\item We are the first to show that models trained independently on distinct, non-overlapping single lives develop a highly consistent geometric understanding. 
\item We curate and analyze  a diverse set of single-life data sources (20 `lives'), including week-long recordings of up to 38 hours from 4 distinct lives, to fuel our exploration into `single-life' learning.
    \item We introduce a metric for measuring the functional similarity of learned visual representations at patch level, providing a new tool for analyzing and comparing independently trained visual models.
    \item We demonstrate that the single-life learning paradigm is viable, showing that these models learn generalizable representations that successfully transfer to downstream geometric tasks in unseen environments.
\end{enumerate}

While the `single life' paradigm in our experiments is only conceptual, its success can revolutionize embodied learning, personalization of models and a shift from diverse scale training to learning in limited data regimes. 
\section{Related Work}
\paragraph{Representation learning from video streams.}
Early explorations into long-term egocentric video captured by one participant, such as the 70-hour KrishnaCam dataset~\cite{krishna-wacv2016}\footnote{Unfortunately, this resource offers low quality videos 10+ years old.}, demonstrated the potential of leveraging the inherent redundancies in one person's life for scene understanding tasks. Inspired by how humans learn from long video streams, some works train vision models from a single continuous video. The most related is DoRA~\cite{venkataramanan2024dora}, which uses a single long video with aggressive data augmentation and IID sampling to learn a semantic representation comparable to ImageNet pre-trained models. DoRA focuses on semantic tasks such as segmentation and detection but did not attempt to learn geometry, or correlate signals from different videos. Other approaches strictly follow the order of video frames, including frameworks that learn with small batch size~\cite{carreira2024bl}, train with multi-tasks~\cite{yan2025stream}, 
and optimizers that tackle the gradient correlation in redundant data~\cite{han25orthogonal}. In contrast, our focus is not on learning from streams, but on the sufficiency of data from a \emph{single life} and on comparing the similarity of representations learned across different lives.
\paragraph{Geometry-focused representation learning.}
Our work is also inspired by recent progress in learning visual representations with strong geometric awareness.
Our training is based on the cross-view completion objective used by CroCo~\cite{weinzaepfel2022croco,weinzaepfel2023croco} and Siamese MAE~\cite{gupta2023siamesemae}, which are adaptations of the Masked Autoencoder~\cite{he2022masked} to a Siamese architecture. This objective has proven effective for learning representations that enable a range of geometric 3D vision tasks~\cite{wang2024dust3r,an2025cross,danier2025depthcues}. While we build on this architecture, we depart from these works by investigating its potential when trained not on diverse, internet-scale data, but on egocentric videos from a single person.
\paragraph{Representation alignment.}
A key question we pose is to what extent do representations learned from different lives exhibit functional similarity. In this, we follow recent studies exploring \textit{representation alignment}, where networks, trained independently on distinct data, converge to similar representations~\cite{kornblith2019cka,bansal2021revisiting}. 
Several works have investigated the similarity of self-supervised representations across modalities~\cite{huh2024platonic,maniparambil2024do,schnaus2025s,zhang2025assessing,jha2025harnessing,tjandrasuwita2025understanding}, notably through the lens of the Platonic Representation Hypothesis~\cite{huh2024platonic}, which posits that models converge to a shared statistical model of reality. We build on this line of research by comparing representations learned independently from different lives. 

\paragraph{Egocentric video understanding.}
Lastly, our work is related to egocentric video understanding, a field driven by large-scale datasets~\cite{Damen2022RESCALING,Ego4D2022CVPR,perrett2025hdepic}. Most prior work has focused on interpreting semantic content for tasks like action recognition and object detection. While some have used egocentric video for geometric tasks such as ego-motion estimation~\cite{krishna-wacv2016,zhou2017unsupervised,mai2023egoloc,EPICFields2023,engel2023projectarianewtool}, the primary goal has been trajectory prediction rather than learning generalizable feature representations. Relatedly, camera estimation efforts from egocentric input~\cite{EPICFields2023,engel2023projectarianewtool}, and accompanying sensors showed that these egocentric datasets  capture indoor walking paths.
In contrast, we use the continuous and redundant nature of a first-person video as a rich, self-supervised signal to investigate a fundamental question: can a generalizable and aligned geometric representation of the world emerge from a single life's experience?

\section{Framework for Single-Life Learning}

\subsection{Formulation: Learning from a Single Life}

The fundamental corpus of data in this paper is a ``life'' which we define as the stream of egocentric video captured by a single individual, denoted $\mathcal{D}_i$ for the $i$-th participant. 
As a continuous stream might not be present, we use the set of videos all captured by the same individual.
We investigate the properties of different `lives' data and of models trained exclusively within the confines of such individual, typically non-overlapping, visual experiences.

Instead of aggregating data to train one unified model, our ``single-life learning paradigm'' trains a distinct model $f_{\theta_i}$ for each life, starting from a random initialization. For each participant $i$ in a set of $n$ individuals, we optimize a separate set of parameters $\theta_i$ \textit{using only their data} $\mathcal{D}_i$:
\begin{equation}
\theta_i^* = \arg\min_{\theta_i} \mathcal{L}(f_{\theta_i}, \mathcal{D}_i).
\end{equation}
This yields a set of $n$ independently trained models, characterized by their optimized parameters $\{\theta_1^*, \theta_2^*, \dots, \theta_n^*\}$. This approach stands in contrast to the dominant learning paradigm, which aggregates all available data from countless sources to learn one general-purpose representation.

Our work focuses on two primary properties of this set of `single-life' models:
\paragraph{(i) Alignment.} We investigate whether models trained on different lives develop a functionally similar geometric understanding. To quantify this, we measure the representational similarity between any two models, ${\theta_i^*}$ and ${\theta_j^*}$ (where $i \neq j$). A high alignment score would suggest that a shared geometric foundation can be learned independently from different individual experiences. We motivate and define  our metric in Section~\ref{sec:alignment}.

\paragraph{(ii) Generalization.} We assess the practical utility of a single-life model ${\theta_i^*}$ by evaluating its ability to generalize to novel data. This is measured by its performance on standard downstream geometric tasks (e.g., depth estimation) with a task-specific head, in order to test whether a model trained on one life can adapt to unseen environments.

\subsection{Model Architecture}

To investigate the single-life paradigm, we require an architecture capable of learning from the geometric signals inherent in an egocentric video stream. We adopt the Cross-View Completion (CroCo) architecture~\citep{weinzaepfel2022croco}, a model recognized for learning robust geometric representations by observing scenes from multiple viewpoints. This objective directly leverages the spatiotemporal structure present in a single life, where diverse views are naturally captured over time. This architectural choice is supported by recent findings; for instance, ZeroCo~\cite{an2025cross} found that cross-view completion models considerably outperform contrastive methods like DINOv2~\cite{oquab2023dinov2} on correspondence tasks, suggesting that this objective ``encodes richer geometric information''. 

The core training principle for CroCo is to learn from pairs of images that depict overlapping scenes. Our key departure is not in the architecture itself, but in the highly constrained data regime from which these pairs are sourced: replacing synthetic and curated massive scale pairs of images in~\citep{weinzaepfel2022croco} with pairs of images sourced from a distinct user's visual experience. The specific strategies for sampling these pairs are detailed in Section~\ref{sec:pairing}.

We use the same architecture as CroCo, a Siamese encoder-decoder transformer. We define an input pair as a \textit{source} image, $x_s$, and a \textit{target} image, $x_t$. Following the masked autoencoder paradigm~\citep{he2022mae,tong2022videomae}, both images are divided into a sequence of non-overlapping patches, $p_s$ and $p_t$. A high ratio of the target image's patches is randomly masked, denoted as $p_t^{\text{masked}}$, leaving a small subset of visible patches, $\tilde{p}_t$.

A weight-sharing ViT encoder, $\mathcal{E}_\theta$, processes the full set of source patches $p_s$ and the visible target patches $\tilde{p}_t$ independently to produce latent feature embeddings:
\begin{equation}
\mathbf{z}_s = \mathcal{E}_\theta(p_s) \quad \text{and} \quad \mathbf{z}_t = \mathcal{E}_\theta(\tilde{p}_t).    
\end{equation}

A transformer decoder $\mathcal{D}_\phi$ reconstructs the masked target patches. The decoder is composed of a series of attention blocks. Within each block, the target features $\mathbf{z}_t$ (along with learned mask tokens) form the query matrix $Q$, while the source features $\mathbf{z}_s$ form the key $K$ and value $V$ matrices. The cross-attention mechanism between $Q$, $K$, and $V$ captures the pairwise similarity between target and source patch tokens which we leverage for our alignment analysis in Section~\ref{sec:alignment}. The decoder uses the output of this attention mechanism to predict the masked target patches, $\hat{p}_t = \mathcal{D}_\phi (\mathbf{z}_s, \mathbf{z}_t)$. The entire model is trained end-to-end by minimizing the Mean Squared Error (MSE) between the reconstructed and original pixels of the masked target patches~\citep{weinzaepfel2022croco}: $\mathcal{L}_{\text{CroCo}} = \sum \lVert \hat{p}_t - p_t^{\text{masked}} \rVert ^2$.

\subsection{Pairing Strategies for Single-life CroCo}\label{sec:pairing}
The effectiveness of cross-view completion is tied to the quality of the image pairs it is trained on. The learning signal for geometric correspondence arises from the 3D spatial \textit{transformations} between two overlapping views of a scene; with no overlap, the task collapses to single-image reconstruction, and with identical views, it becomes trivial.

Previous work like CroCo~\cite{weinzaepfel2022croco} used this training strategy on curated pairs from static synthetic scenes, simplifying the task to modeling only camera motion. 
Different from these works, we aim to sample frames from egocentric data, capturing the natural movement of a person around an environment.
This offers a previously unexplored resource for multi-view geometry training.
However, real-world egocentric video introduces two key challenges. First, the viewpoints are not sampled uniformly but are limited to a \textbf{functional subset}, captured as a person traverses an environment or performs an activity. Second, scenes are inherently \textbf{dynamic}, with moving objects and changing lighting. The model must therefore learn to distinguish geometric viewpoint shifts from appearance (i.e. dynamic) changes. Our work extends the cross-view completion paradigm to this more realistic and complex setting. We use two signals to generate pairs for learning.

\paragraph{Temporal pairing.}
Our first signal, \textit{temporal proximity}, leverages the natural continuity of video. The assumption is that most temporally close frames have a non-trivial viewpoint overlap. This strategy is similar to~\cite{gupta2023siamesemae}, which sampled frames from diverse web clips to learn short-term correspondence. 
In contrast, we are the first to apply this pairing strategy to long-form egocentric video.
This versatile method requires no ground truth beyond the video itself. 

\paragraph{Spatial pairing.}
Our second signal is inspired by proprioception — a living being's innate sense of their body's position and orientation in space. This sense allows humans to recognize that they are observing the same scene from a different viewpoint, even after a significant time gap. To model this, our spatial pairing strategy uses camera poses and 3D point clouds (when available) to identify such image pairs. We search for pairs of frames that have a high degree of overlap, measured by computing the Jaccard index between the visible 3D point clouds of two frames. 

We analyze models trained with temporal pairs, spatial pairs, and their union. Intuitively, their combination should be most powerful, mirroring how humans integrate continuous motion with the recognition of significant viewpoint changes. Our findings confirm this: models trained on the union of both strategies (spatial and temporal) consistently learn the most  generalizable representations.

\subsection{Patch-Level Alignment Metric of Models}\label{sec:alignment}

As highlighted above, we train a separate model  $\theta_i^*$ on data derived from a video capturing a distinct individual life $i$. Given $n$ such lives, we therefore obtain $n$ distinct trained models, which we denote as $\{\theta_1^*, \theta_2^*, \dots, \theta_n^*\}$. 

\begin{figure}
    \centering
    \includegraphics[width=\linewidth]{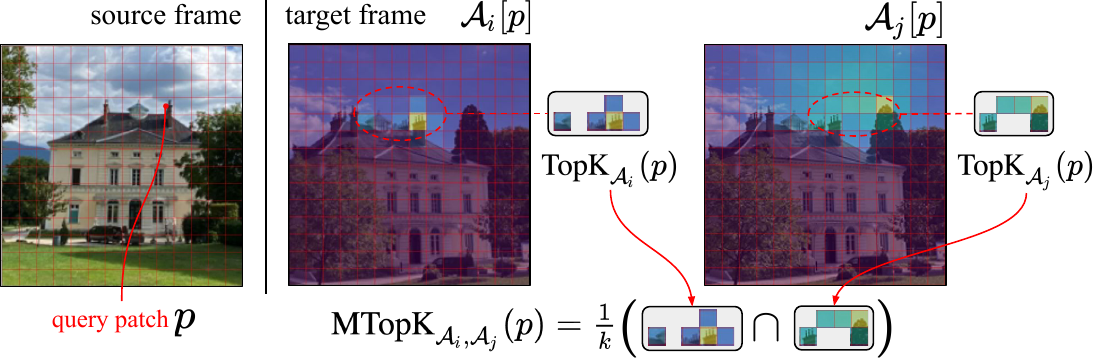}
    \caption{\textbf{Correspondence Alignment Score (CAS).} Given a test pair of  images, we extract cross-attention maps, $\mathcal{A}_i$ and $\mathcal{A}_j$, from independently trained models $\theta^*_i$ and $\theta^*_j$. For each query patch in the source image, we identify the top-K most attended-to patches in the target  by both models, and compute their intersection.
    }
    \label{fig:CAS}
\end{figure}

Our goal is to assign a \textit{similarity score} to a pair of models $(\theta_i^*, \theta_j^*)$, without relying on additional training. To derive our similarity metric we extend the robust mutual $k$-nearest neighbor score introduced in 
\cite{huh2024platonic}, to a new metric, which evaluates how different models relate images on the patch level.
We first construct an evaluation set by sampling pairs of images, with a known overlap, from a withheld test set. Given a test pair of source and target images $\{x_s, x_t\}$ containing $N$ patches each, following~\citep{an2025cross}, we construct a cross-attention map for each trained model, by aggregating the decoder attention map $A = \frac{1}{d} \sum_{b=1}^{d}{ \mathbf{q}_b^\top \mathbf{k}_b}$ across all $d$ decoder blocks. Let $A_i, A_j \in \mathbb{R}^{{N \times N}}$ be the aggregated cross-attention maps from models $\theta_i^*$ and $\theta_j^*$ respectively. For any patch $p \in x_s$, we let $\text{TopK}_{A_{i}}(p)$ be the set of $k$ patches in the target image with the highest cross-attention score to $p$ according to $A_{i}$. The mutual top $k$ correspondence measure $\text{MTopK}_{A_i, A_j}(p)$ denotes the fraction of mutual correspondences of $p$, according to $A_i$ and $A_j$: $\text{MTopK}_{A_i, A_j} (p) = \frac{1}{k} | \text{TopK}_{A_{i}}(p) \cap \text{TopK}_{A_{j}}(p)|.$
 
Our \textbf{Correspondence Alignment Score} (denoted as \textbf{CAS}), visualised in Fig.~\ref{fig:CAS}, measures the similarity between two models $\{\theta_{i}^{*},\theta_{j}^{*}\}$, and is then defined as follows:
\begin{align}
\text{CAS}(\theta_{i}^{*},\! \theta_{j}^{*}) \! = \! \frac{1}{|\mathcal{T}|} \sum_{(x_s, x_t) \in \mathcal{T}} \frac{1}{N}  \sum_{p=1}^{N} \text{MTopK}_{A_i,  A_j} (p).
\end{align}
Here, $\mathcal{T}$ is a test set of image pairs $\{(x_s, x_t),\dots\}$ and $N$ is the total number of patches in the source image. 

CAS ranges between 0 (no alignment) and 1 (perfect alignment), and inherits key properties from the $k$-nn metric from \cite{huh2024platonic}, including commutativity, robustness to changes in attention \emph{values}, and being training-free. Importantly, unlike the cross-model similarity scores explored in previous works \cite{huh2024platonic, kornblith2019cka}, which treat entire images as instances, CAS is sensitive to the \textit{patchwise} relations across images induced by different models, and is thus more adapted for measuring geometric or localized model alignment.

\section{Single-Lives Datasets and Control Datasets}

\begin{figure*}[t]
    \centering
    \includegraphics[width=\textwidth]{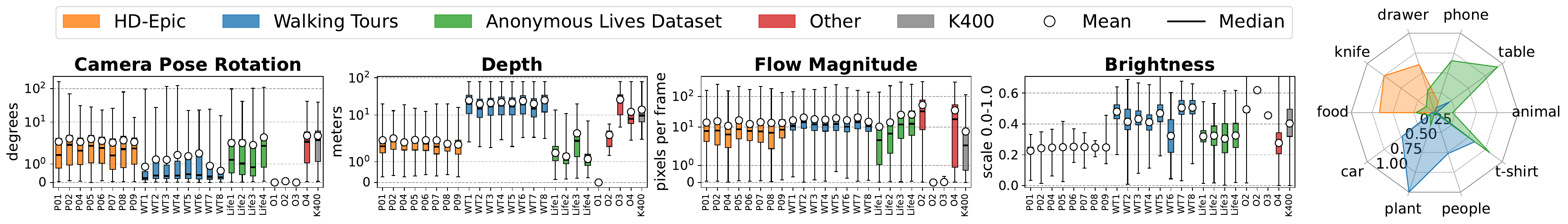}
    \caption{Properties of every `life' used for training (colored by dataset). Left: distribution of continuous variables per life, shown in box plot. Right: relative count of detected objects aggregated per dataset, shown in spider plot. See \app~for full details.\vspace{-2mm}}
    \label{fig:properties_of_lives}
\end{figure*}

We use three sources of egocentric video for single-life training -- two public and one private dataset. 
In total, we train 20 models from single lives and 5 models that contain a comparable amount of data, but do not represent a `life.' 
\paragraph{HD-Epic~\cite{perrett2025hdepic} (8 single-room indoor lives).} We use the public dataset collected from 9 kitchens around the UK.
While the dataset is formed of 156 videos, we restructure these based on the participant/environment to treat videos from each participant as a `life'. 
The dataset is collected using Meta's Aria glasses~\cite{engel2023projectarianewtool} -- a multi-sensored research platform.
As a result, the dataset has ground truth camera poses calibrated for each environment to metric depth.
We choose P03 for `testing', and train 8 models on participants, each sourced from an average of 4 hours of video.

\paragraph{Walking Tours (WT)~\cite{venkataramanan2024dora} (8 outdoor lives).} To represent outdoor experiences, we use the Walking Tours dataset, which contains hour-long, single-shot videos of individuals walking through various urban environments. Originally introduced in~\cite{venkataramanan2024dora}, it includes videos from 9 distinct cities such as Amsterdam, Bangkok, and Venice, each serving as a unique `life'\footnote{We exclude the Wildlife video released as part of WT~\cite{venkataramanan2024dora} as it is neither egocentric nor continuous.}. For brevity, we rename the videos as WT1 to WT9 and hold out the Istanbul video (WT9) for testing and train a separate model on each of the other 8 city videos. 

\paragraph{\cl~(ALD) \cite{adl} (4 long-term lives).} 
We use a GDM dataset\footnote{Data is collected with participant consent for usage in research.} collected in a similar manner to other public egocentric datasets~\cite{Ego4D2022CVPR,Damen2022RESCALING} but for a longer-term.
Participants collect egocentric videos over one week each providing at least 30 hours of recordings (max 38 hours).
Similar to prior work, the collections are unscripted using GoPro devices where participants are not given instructions of what to carry out, allowing them to capture their natural daily activities.
Unlike HD-Epic, which is restricted to kitchens, these videos capture larger indoor and outdoor environments.
The dataset contains 5 lives which we name as Life1 to Life5. We reserve Life5 for `testing'. 

\paragraph{Other `non-life' videos.} To isolate the importance of the egocentric signal, we use four `Non-life' videos as control. These continuous videos lack the characteristic motion and perspective of a human experience. They are each one hour in duration: (\textbf{O1}) a screen recording of a \LaTeX tutorial; (\textbf{O2})~a fixed-camera view of a static indoor room; (\textbf{O3}) a surveillance camera observing a street view with moving objects; (\textbf{O4}) a recording of the video game `Minecraft'.

Moreover, we train one model on diverse data, a subset of Kinetics-400 (\textbf{K400})~\cite{kay2017kinetics}. K400 is a large-scale action recognition dataset containing hundreds of thousands of short, 10 second distinct video clips. By training on size-matched subsets of 30 hours, we can directly compare learning from a single life versus learning from a diverse collection of unrelated videos/scenes.

Figure~\ref{fig:properties_of_lives} compares key properties relevant for geometry-oriented visual learning across these datasets (details in \app~\ref{appendix:dataset}).
Interestingly, the relative camera pose rotation shows that indoor environments exhibit more head rotation, compared to the task of just `walking' in WT. Conversely, outdoor environments allow observing objects at significantly further depth. All lives share similar flow magnitudes and spread. Single-indoor rooms exhibit low variations in brightness, and kitchens seem to significantly be darker than multi-indoor and outdoor environments. Outdoors allows capturing more people and plants compared to single-person recordings in HD-Epic where these are not present. HD-Epic is predominantly food and the means to prepare it, which has some presence in ALD and almost no presence in WT. ALD also shows significant usage of phones and pets compared to the other lives.
In contrast, some properties are undefined (e.g. no flow measured on \textbf{O2, O3} or depth in~\textbf{O1}). 

\section{Experiments}

Our experiments are designed to measure two key properties of our single-life training paradigm: representation alignment and generalization to downstream tasks.

\subsection{Evaluation}\label{sec:evaluation}
\paragraph{Representation alignment.} To compare models trained independently on different lives, we compute the Correspondence Alignment Score (CAS) (as defined in Sec.~\ref{sec:alignment}). We use two protocols: comparing each single-life model to an officially released pretrained CroCo checkpoint~\cite{an2025cross},
which is trained on diverse image pairs from synthetic and real data, with $\text{CAS}(\theta^{*}_i, \theta^{*}_{\text{CroCo}})$, and comparing any pair of single-life models pairwise 
($\text{CAS}(\theta^{*}_i, \theta^{*}_{j})$).
For each held-out `test life' from each dataset (WT9 for Walking Tours, P03 for HD-Epic, Life5 for ALD), we uniformly sample 1,000 image pairs and combine these to form a single test set of 3,000 pairs to compute the CAS metric.

\paragraph{Generalization.} We evaluate generalization on depth estimation and correspondence matching.
\paragraph{\textit{Monocular Depth Estimation:}} Following common practice~\cite{weinzaepfel2022croco,carreira2024scaling}, we evaluate on two standard benchmarks. \textbf{NYU-Depth-V2}~\citep{silberman2012indoor} is a widely used indoor scene dataset with 795-images in the train set, and evaluating the performance on the official 654-image test set and reporting the $\delta_1$ accuracy (the percentage of pixels with an error ratio below $1.25$). \textbf{ScanNet}~\citep{dai2017scannet} is a large-scale 3D reconstruction dataset with 1201 and 312 videos in training and validation sets that provide a more challenging generalization test due to its wide variety of camera poses and environments. We report mean of the absolute
relative error (AbsRel)~\citep{ranftl2021vision} on the validation set which is computed as $|d^* - d|/(d + \epsilon)$ where $d^*$ is the predicted depth values and $d$ is the ground truth depth.
We use \emph{attentive probing} on downstream depth tasks unless otherwise mentioned. Specifically, we follow the standard protocol used in \citep{carreira2024scaling} for which we train a lightweight single attention block as the readout module on top of a frozen encoder. Full-finetuning results are in the~\app.

\paragraph{ \textit{Zero-Shot Correspondence:}} To assess the robustness of our learned local descriptors under viewpoint and illumination changes, we evaluate on the \textbf{HPatches}~\citep{balntas2017hpatches}, a benchmark of image pairs annotated with optical flow.
Following the protocol of~\citep{an2025cross}, we use the model's cross-attention to compute dense similarity between features in a zero-shot manner. We report the mean Average End-Point Error~(AEPE) across all pairs which is the average Euclidean distance between predicted and the ground truth optical flow. This approach directly probes the representation's geometric matching capabilities without any fine-tuning.

\subsection{Results}
This section presents our experimental findings, structured as a series of questions and answers.

\begin{qabox}{When does alignment to standard models emerge?}
We observe a critical duration ranging from 30 mins to 2 hours when alignment emerges. 
\end{qabox}

\begin{figure}[t]
    \centering
    \includegraphics[width=0.4\textwidth]{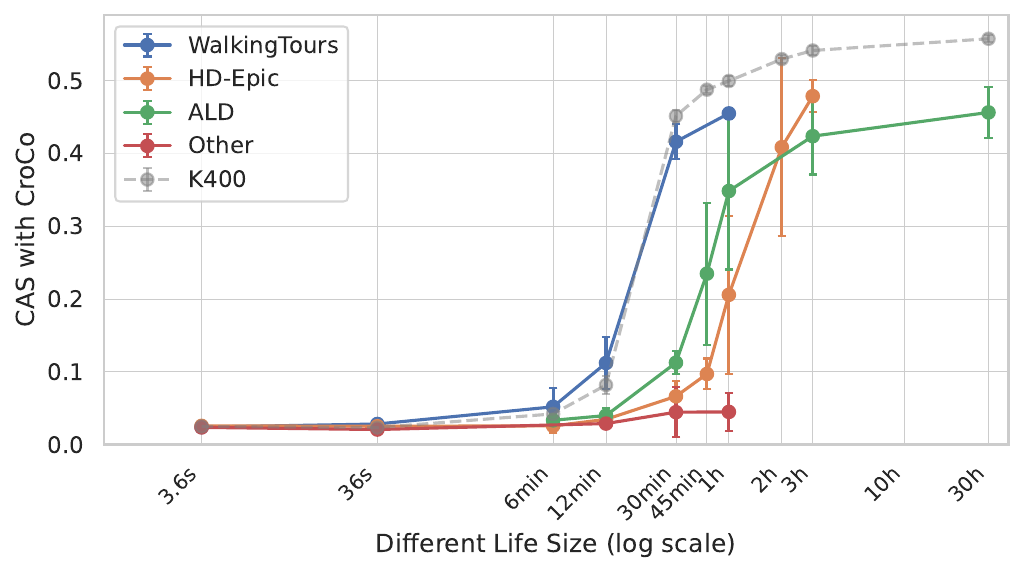}
    \caption{Effect of `life-size' measured by CAS with CroCo. For WalkingTours, HD-Epic and ALD, the error bars (standard deviation) are obtained from multiple lives. For K400, the tiny error bars are from different random seed when creating subsets. 
    }
    \label{fig:scale_croco}
\end{figure}

We address our central hypothesis regarding representation alignment. Figure~\ref{fig:scale_croco} explores the effect of `life-size' on alignment measured by CAS comparing a single-life model to the CroCo checkpoint - $\text{CAS}(\theta^*, \theta_{\text{CroCo}})$. 
By comparing to the public checkpoint of $\theta_{\text{CroCo}}$, we effectively compare single-life models to a model trained with diverse web-scale data.
We observe that a critical duration — ranging from 30 minutes for Walking Tours, to 1 hour for ALD and 2 hours for HD-Epic — is required for strong geometric alignment to emerge. 
We hypothesise that as CroCo is mostly trained on outdoor scenes, WT aligns to it faster.
This provides direct evidence for our Shared World hypothesis introduced in Sec~\ref{sec:intro}: as models are exposed to more data reflecting the world's shared physics, their functional representations become more similar. 

\begin{qabox}{Do models trained on non-life videos align?}
`Non-life' control videos fail to align. Single-life models are not far behind diverse web data on the same scale. 
\end{qabox}

Also in Fig.~\ref{fig:scale_croco}, we show that the alignment score approaches the upper bound set by the highly diverse K400 dataset.
In contrast, models trained on our `non-life' control videos achieve near-zero alignment with the CroCo baseline. This result demonstrates that merely training on a long video is insufficient; the egocentric perspective, with its characteristic motion and interaction, is the essential ingredient for learning the shared world. While our longest single-life videos are still orders of magnitude shorter than a human's complete visual experience, the clear scaling trend suggests that training on even larger-scale single-life datasets is a promising direction for future work to further close the gap with diverse-data baselines.

\begin{figure}[t]
    \centering
    \includegraphics[width=0.95\linewidth]{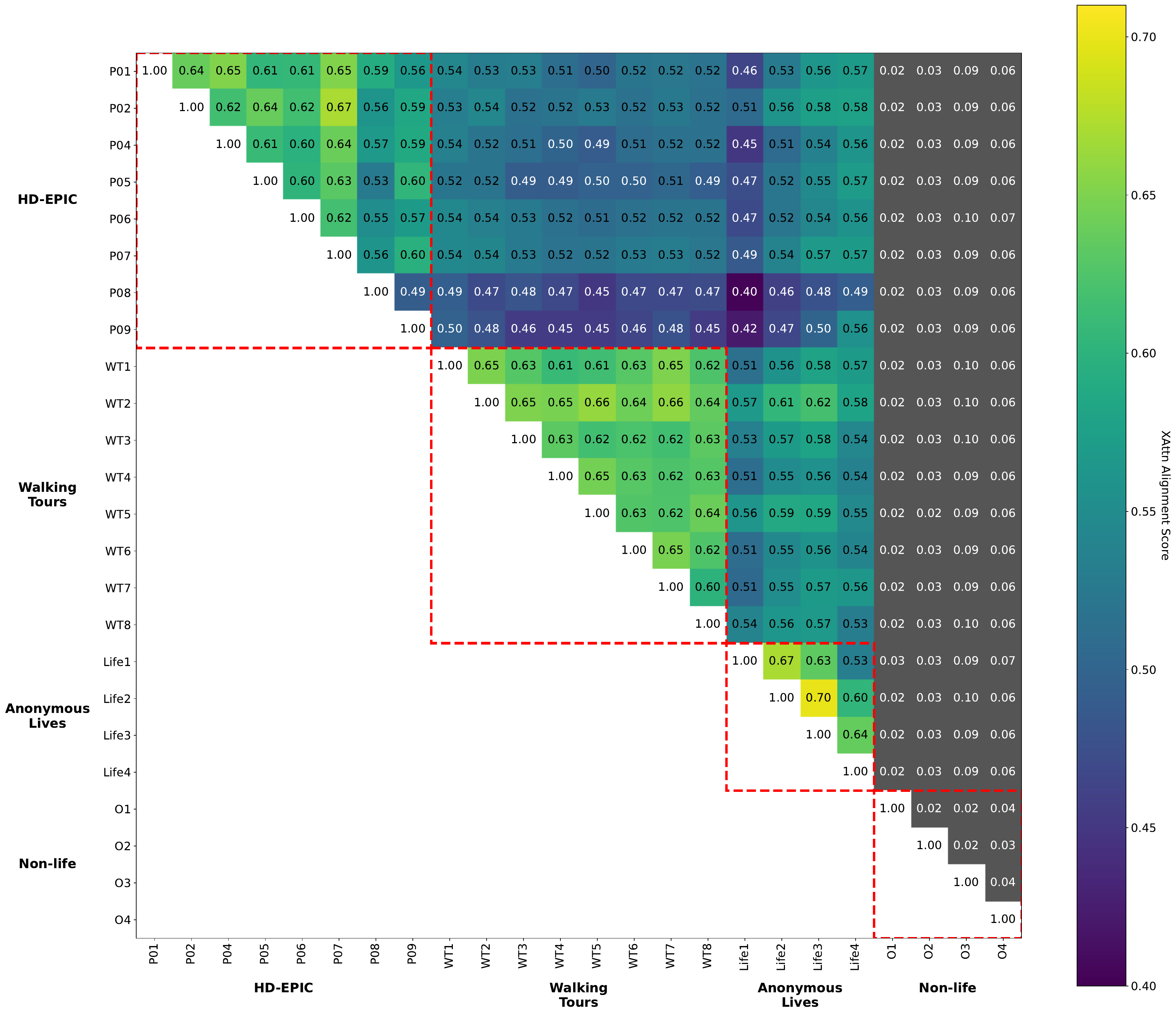}
    \caption{CAS score matrix comparing models trained on individual lives. A higher score signifies a stronger alignment. Values outside the primary range are represented with colors beyond the colormap (values below 0.4 are gray and above 0.71 are white).}
    \label{fig:cas_matrix_a}
\end{figure}

\begin{qabox}{What about alignment within Single-Life models?}
Single-life models align with increasing degrees depending on the similarity of their captured environments.
Models from similar environments (e.g., indoor kitchens vs.\ outdoor tours) form distinct alignment clusters.
\end{qabox}

\begin{figure}[t]
    \centering
    \includegraphics[width=0.83\linewidth]{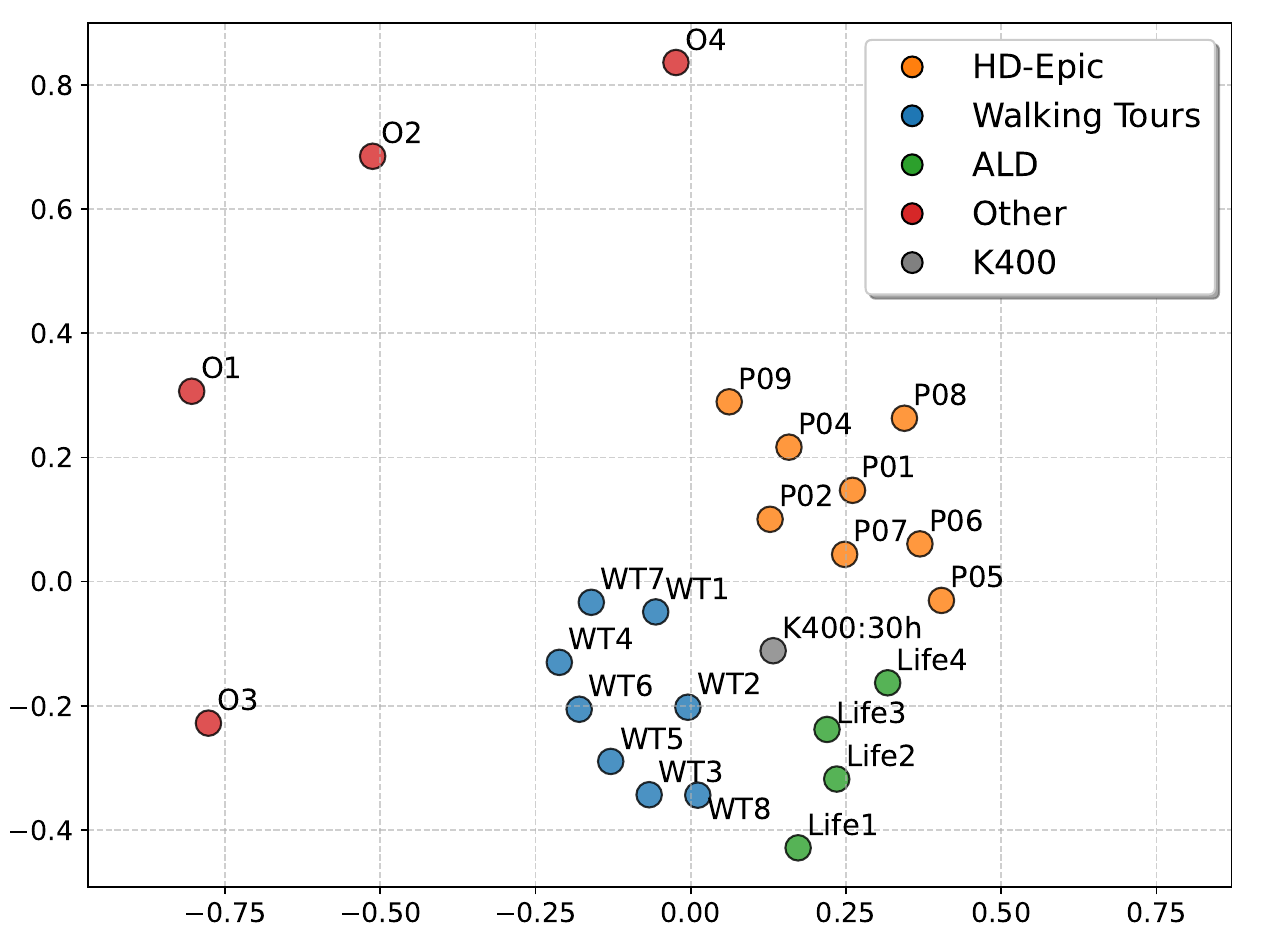}
    \caption{A 2D MDS visualization of the models, using the CAS score from Fig.~\ref{fig:cas_matrix_a} as the similarity metric. Observe the strong clustering of ``successful'' models and the central position of the model trained on a general dataset (K400 30h).}
    \label{fig:cas_matrix_b}
\end{figure}

Figure~\ref{fig:cas_matrix_a} and~\ref{fig:cas_matrix_b}  visualize the cross-model alignment -- measured by $\text{CAS}(\theta^*_j, \theta^*_{j})$ in two ways -- through a confusion matrix and a 2D multidimensional scaling (MDS) projection~\cite{carroll1998multidimensional} -- revealing several key insights. First, the block-diagonal structure in Fig.~\ref{fig:cas_matrix_a} shows that models trained on lives from the \textit{same} dataset (e.g., all kitchens in HD-Epic) align more strongly with each other than with models from \textit{different} datasets (e.g., kitchens vs. outdoor tours). This is further visualized in the MDS embedding (Fig.~\ref{fig:cas_matrix_b}), where models cluster by dataset. This suggests that while all lives learn a shared foundation of geometry, domain-specific properties (as shown in our analysis in Fig.~\ref{fig:properties_of_lives}) leave a distinct imprint on the learned representations. 
Notably, a corresponding 30 hours of K400 seems to average the various life types appearing as the center between these.\cite{he2022mae}

\begin{figure*}[t]
\centering
\begin{subfigure}[b]{0.32\textwidth}
    \centering
    \includegraphics[width=\linewidth]{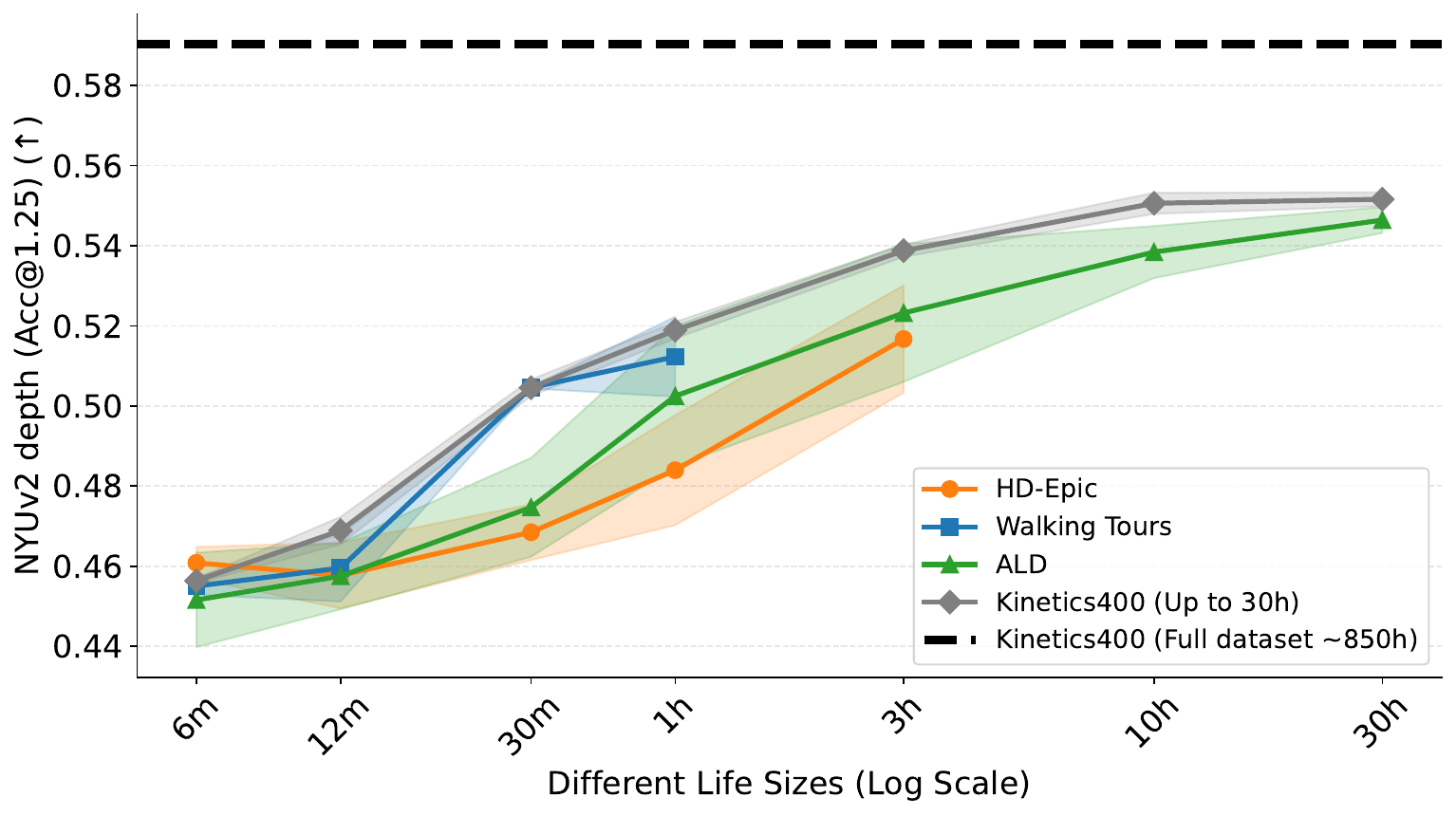}
    \label{fig:nyu_depth_results_avg_log_line}
\end{subfigure}
\begin{subfigure}[b]{0.32\textwidth}
    \centering
    \includegraphics[width=\linewidth]{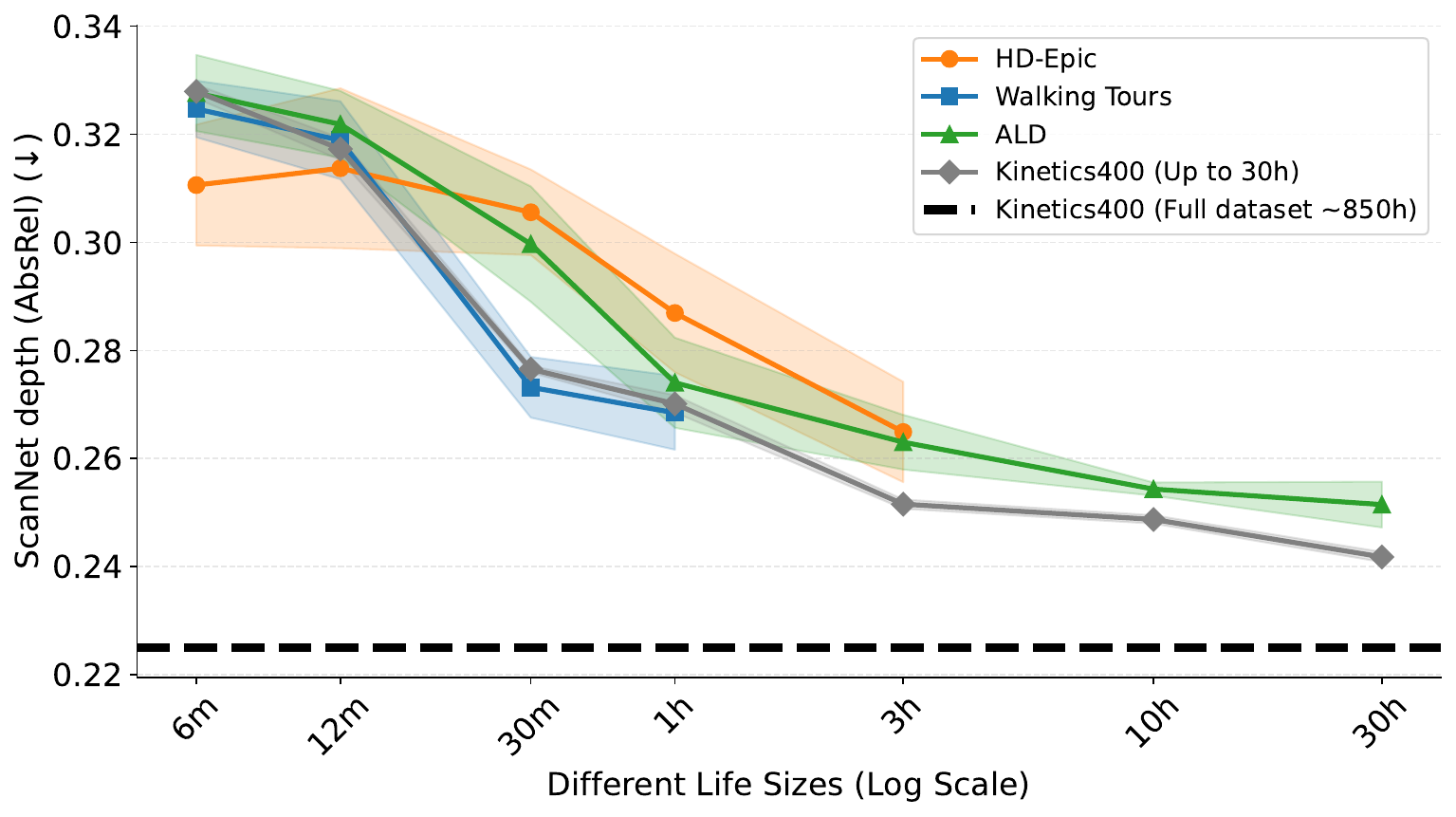}
    \label{fig:scaling_depth_results_avg_log_line}
\end{subfigure}
\begin{subfigure}[b]{0.32\textwidth}
    \centering
    \includegraphics[width=\linewidth]{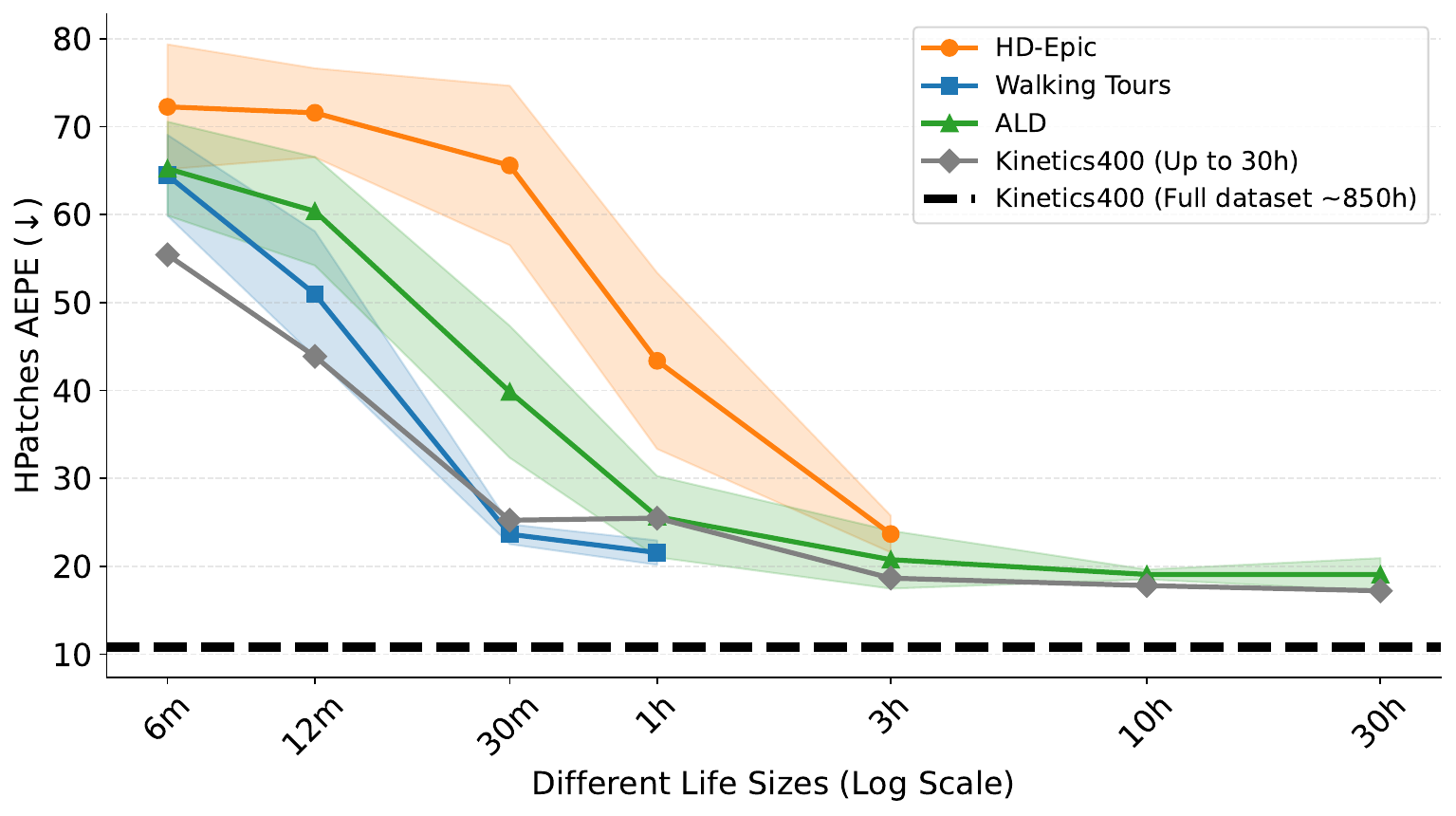}
    \label{fig:hpatches_aepe_results_avg_log_line}

\end{subfigure}
\vspace{-3mm}
\caption{
    {Generalization performance scales robustly with single-life data duration.}
     We evaluate models trained on single lives of increasing duration (colored lines show std. dev. across lives) against a size-matched diverse baseline (K400, gray line) on three downstream tasks: \textbf{(Left)} NYU-Depth-v2 depth estimation, \textbf{(Middle)} ScanNet depth estimation, and \textbf{(Right)} HPatches zero-shot correspondence. While diverse data offers an initial advantage, single-life performance scales remarkably well. Notably, at $\sim$30h, models trained on single lives (especially ALD) become highly competitive, matching or even surpassing the size-matched K400 baseline on several tasks.
    }
\label{fig:downstream_scale}
\end{figure*}
\begin{figure}[t]
    \centering
    \begin{subfigure}{0.48\linewidth}
        \includegraphics[width=\linewidth]{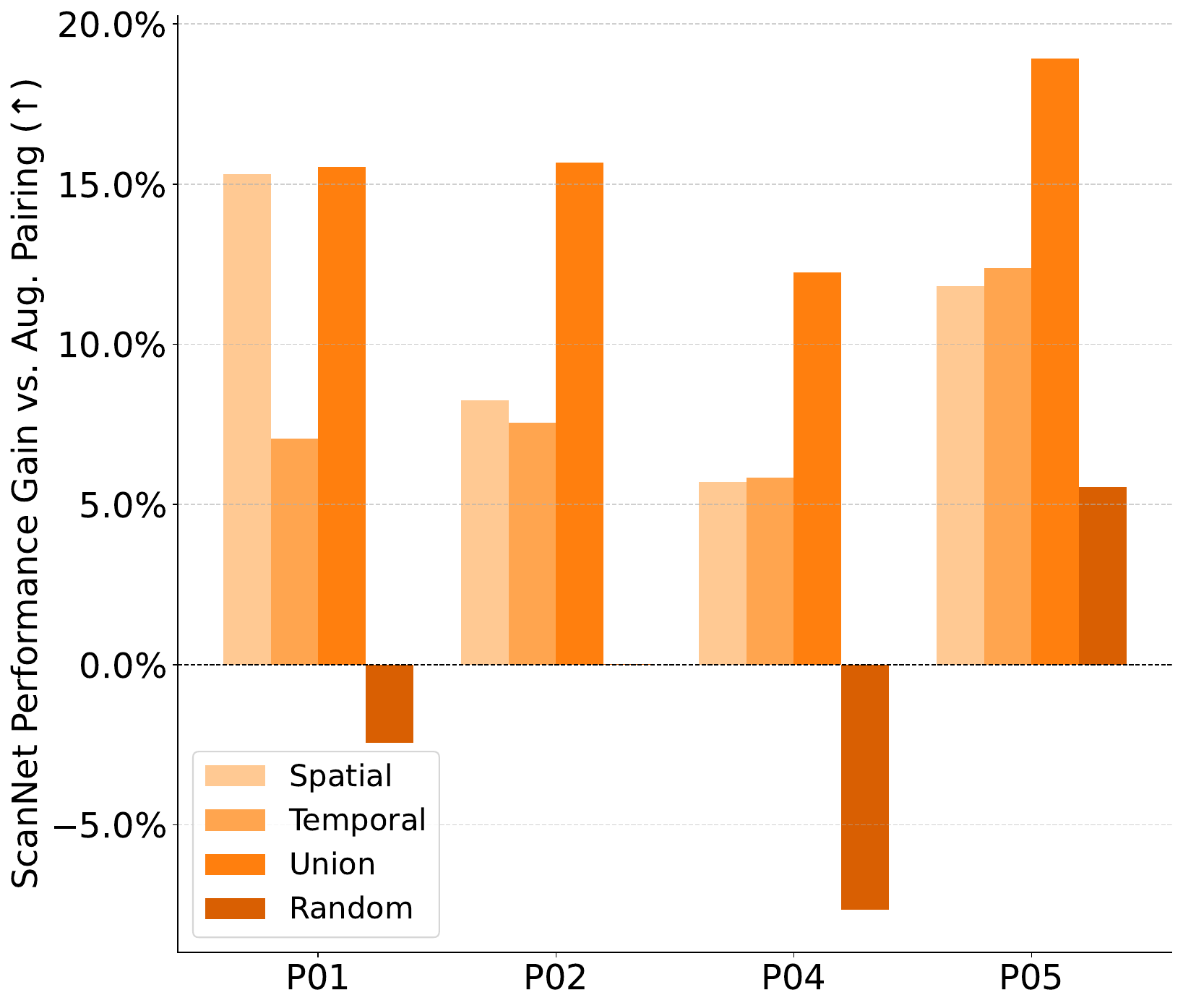}
        \label{fig:scannet_gains}
    \end{subfigure}
    \hfill
    \begin{subfigure}{0.48\linewidth}
        \includegraphics[width=\linewidth]{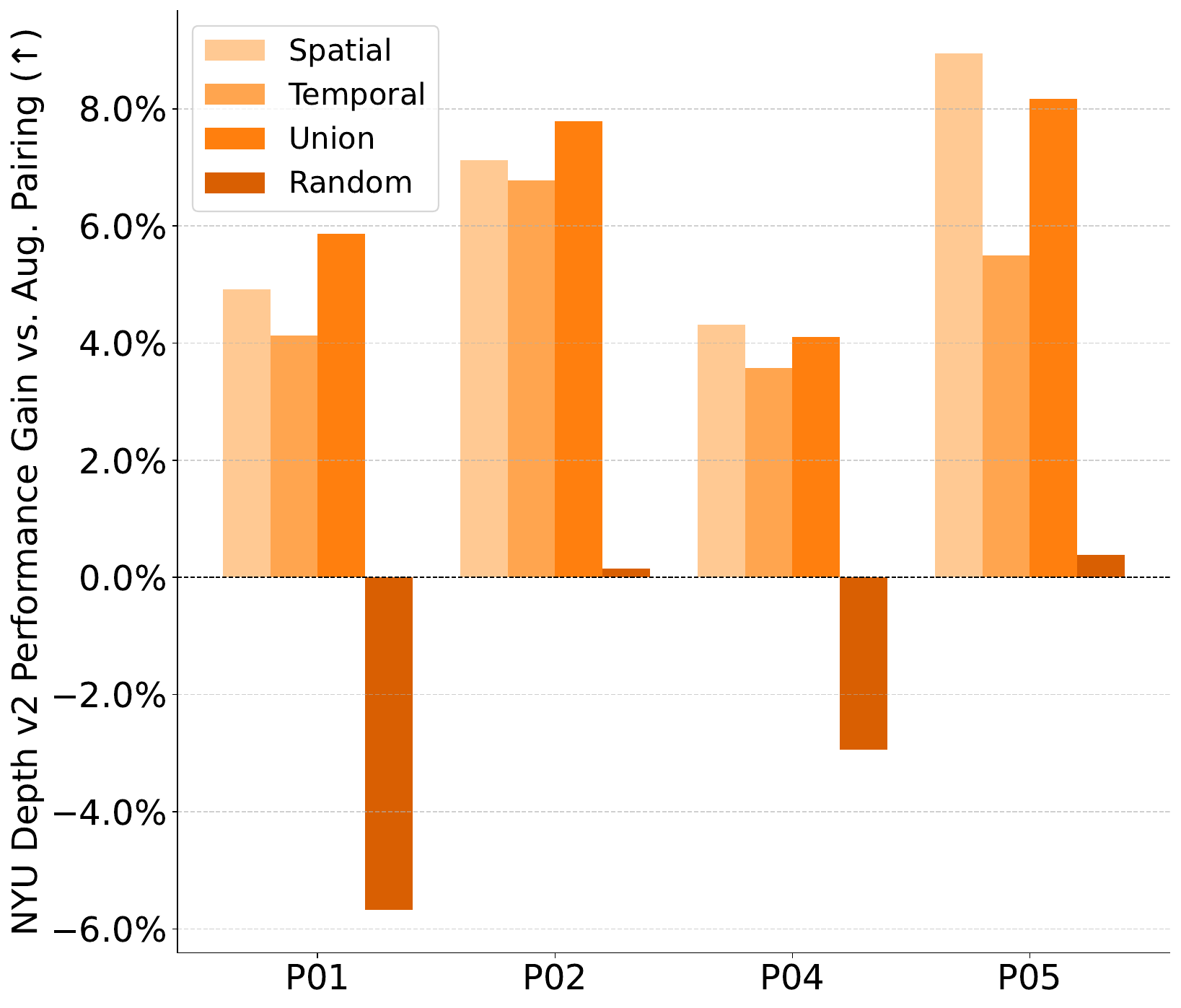}
        \label{fig:nyu_gains}
    \end{subfigure}
    \caption{Relative performance gains of single-life learning on HD-Epic across different pairing strategies. 
    Models are evaluated on ScanNet (left) and NYU-Depth-v2 (right) for monocular depth estimation task by attentive probing. 
    Results for the first four lives are shown here, with full results in the~\app.\vspace{-2mm}}
    \label{fig:hdepic_geometry}
\end{figure}

\begin{qabox}{Can Single-Life models generalize?}
Yes. They learn generalizable representations that transfer to geometry-oriented downstream tasks, with performance scaling robustly with the amount of data.
\end{qabox}

Having established that alignment emerges, we now assess if these representations generalize. Figure~\ref{fig:downstream_scale} plots performance on three geometric tasks for models trained on single lives of increasing duration, revealing an insightful relationship between single-life data and diverse data. While diverse K400 data often provides a strong initial signal, leading to better performance with very short training durations (e.g., under 30 minutes), the performance of single-life models scales very well. For longer-duration lives, particularly the data from the long ALD dataset, the performance gap narrows significantly and in some cases is entirely eliminated. For instance, on NYU-Depth-v2 (Left) and HPatches (Right), the average of the models trained on lives from the ALD dataset
nearly matches or even surpasses the performance of the size-matched K400 baseline at the 30-hour mark. On the diverse ScanNet benchmark~(Middle), the K400 data shows a distinct advantage, although the Walking Tours model remains on par with it for up to one hour of training data.

This establishes two key findings. First, single-life data is a powerful and highly effective source for learning robust, transferable geometric priors. Second, it suggests that while diverse data may accelerate learning, the rich structure of a single long-term life can learn a similarly effective, and perhaps more nuanced, model of the world's geometry. Furthermore, the strong correlation between the scaling trends in alignment (Fig.~\ref{fig:scale_croco}) and generalization (Fig.~\ref{fig:downstream_scale}) provides evidence for our Shared World Hypothesis: models that are well-aligned also tend to generalize effectively to downstream tasks. We show the results for zero-shot correspondence on HPatches in the \app.
\begin{qabox}{How does the single-life generalization compare to the diverse data models?}
Performance is comparable to data of the same size.
\end{qabox}
Comparing against a strong baseline trained on large scale and diverse (horizontal line in Fig.~\ref{fig:downstream_scale}) highlights the effectiveness our paradigm. While models trained on the full K400 ($\sim$850h) dataset set a high-performance ceiling, a model trained on just 30 hours of a single ALD life is already competitive with one trained on 30 hours of the diverse K400 dataset. This suggests that the dense, structured signal within a single life can be a more viable
source for learning geometric priors than an equivalent amount of diverse data.
\begin{qabox}{Which pairing strategy is most effective?}
When sampling pairs of images within a life, the union of spatial and temporal pairing achieves best performance, but the simpler readily available temporal strategy is remarkably effective on its own.
\end{qabox}
We start with comparing different pairing strategies on HD-Epic. Figure~\ref{fig:hdepic_geometry} compares three matching strategies: spatial pairing, temporal pairing, and the union of the two strategies. 
For direct comparison, we report gains compared to a model trained with \textbf{Augmented Pairing}, i.e. pairs generated by 2D transformations of the same image. 
Additionally, we report performance on \textbf{Random Pairing}; a model trained on randomly sampled image pairs from the same video, which might lack correspondence.

We monitor the generalization performance on downstream depth estimation tasks, by attentively probing the trained encoder on (a) ScanNet and (b) NYU-Depth-V2.
Note that we report relative performance difference, and set `Augmented Pairing' as $0\%$ horizontal line. 
The union of spatial and temporal pairs obtains the highest gains.
This observation matches our intuition that humans use both spatial and temporal cues to obtain geometric representation from their surroundings.

\begin{figure}[t]
    \centering
    \begin{subfigure}{0.45\linewidth}
     \centering
\includegraphics[width=\linewidth]{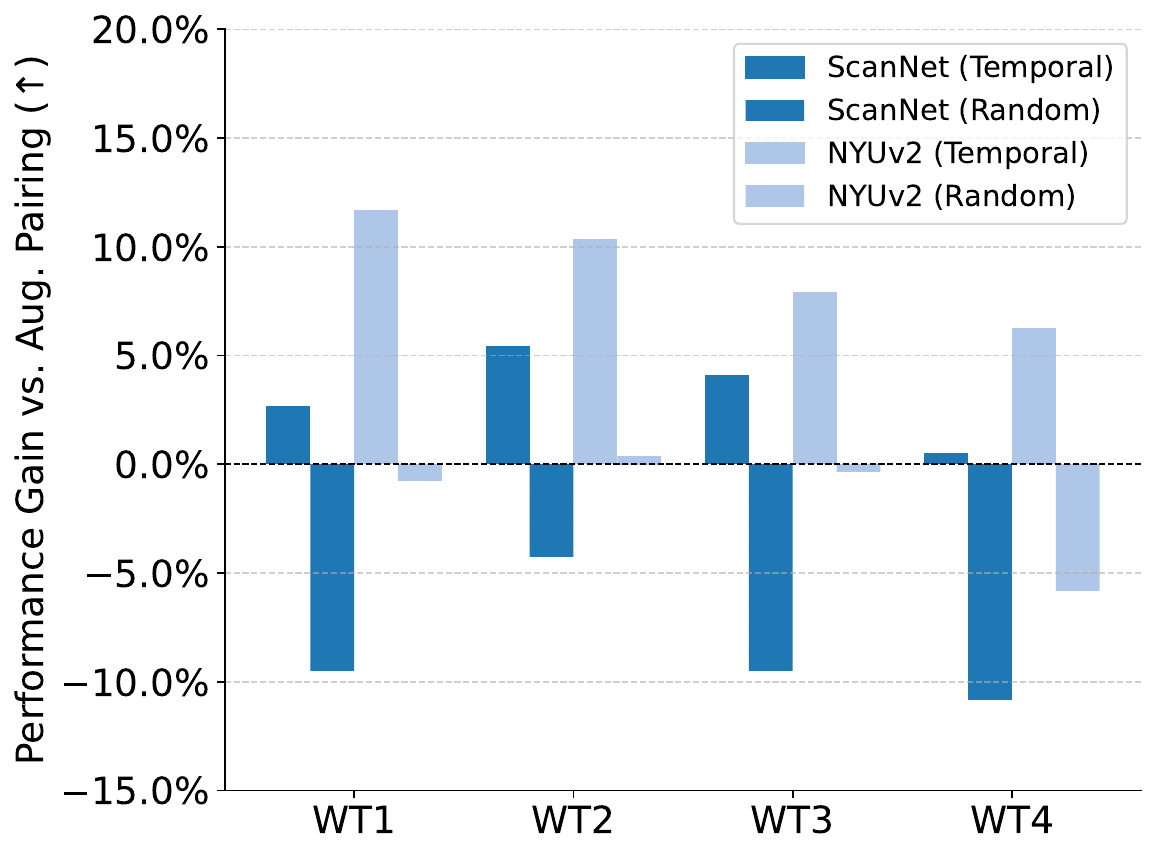}
    \end{subfigure}
    \hfill
    \begin{subfigure}{0.45\linewidth}
    \centering\includegraphics[width=\linewidth]{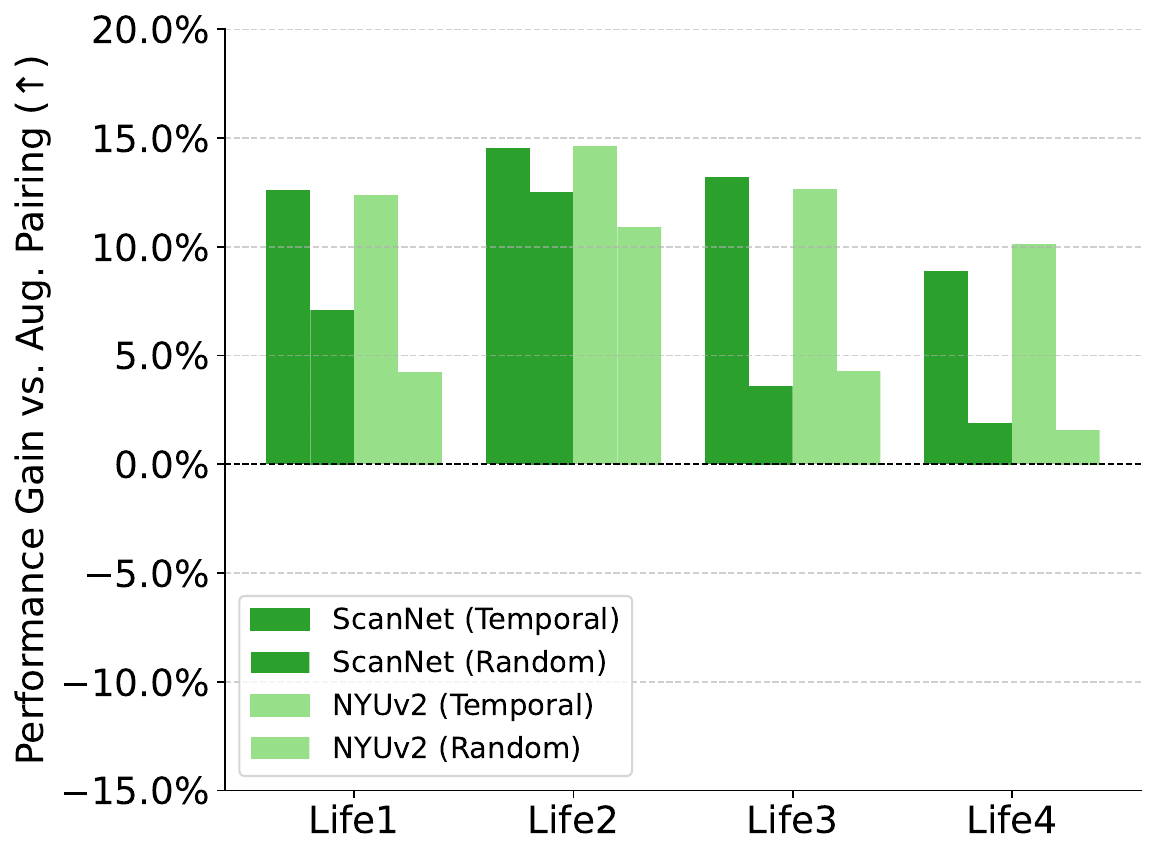}
    \end{subfigure}    
    \caption{Relative performance gains of single-life learning on Walking Tours (left) and \cl~(right) with temporal changes followed by attentive probing on depth estimation on ScanNet and NYU-Depth-v2. Results for the first four cities/lives are shown here, with full results in the~\app.}
    \label{fig:geometry_wt_ald}
\end{figure}
Figure~\ref{fig:geometry_wt_ald} also shows depth estimation gains relative to the Augmented Pairing baseline ($0\%$ line). For Walking Tours, models trained with Temporal Pairing consistently outperform both the Augmented and Random Pairing baselines across all evaluated lives on both ScanNet and NYU-Depth-v2. On ALD, while Temporal Pairing also remains the top performer, Random Pairing notably surpasses the Augmented Pairing baseline. 
We speculate that Random Pairing might still find views with viewpoint overlap as one might remain in the same spot/environment for significant chunks of the video.

\begin{qabox}{Can the single-life paradigm performance extend beyond the CroCo architecture?}
Yes. Single-life training is effective for other learning objectives. We train DINOv2 on single lives, where it also learns generalizable geometric representations.
\end{qabox}
Our work focuses on the CroCo architecture as a commonly accepted paradigm for learning from multiple viewpoints.
An interesting question is whether our findings can be more broadly applicable. To investigate this, we trained a DINOv2 model~\cite{oquab2023dinov2}, which uses a contrastive regime that combines single-image masking with distillation. On the NYU-Depth-v2 benchmark, the DINOv2 models pretrained on Walking Tours cities, achieve a mean $\delta_1$ accuracy of $0.562\pm0.026$) compared to CroCo model's ($0.538\pm0.001)$. The order of the architectures though shifts for other downstream tasks. On the zero-shot HPatches correspondence, CroCo achievs AEPE of $19.4\pm0.38$ vs DINOv2's $21.2\pm4.16$. As detailed in the \app  ~(Section~\ref{appendix:results}), the DINOv2 models also learn effective representations that successfully generalize. This strengthens our contribution of single-life learning, as not being specific to a particular  training strategy.
The question of which self-supervised architecture best learns geometric representations remains an exciting direction for future research.

\section{Conclusion}
This paper introduced the ``single-life'' learning paradigm. Our central premise, the Shared World Hypothesis, posited that the universal physics governing our world would guide independently trained models toward a common geometric understanding. 
Our results provide evidence for this hypothesis.
First, we curate a benchmark of 20 Single-Life datasets, spanning indoor and outdoor scenes, with up to 38 hours of data.
Second, using a novel cross-attention metric, we demonstrated that models trained on distinct lives converge to a remarkably aligned functional representation, a phenomenon that emerges and strengthens with as little as 30-60 minutes of experience. Third, we showed that models trained on Single Lives transfer effectively to downstream geometric tasks such as depth estimation. Fourth, we show that these models are competitive
with an equivalent amount of diverse web data from K400.
Our results establish that the egocentric stream of a single life is a powerful source of supervision for learning geometric priors. Future work should explore the scalability of this paradigm and its potential for learning richer \textit{semantic} representations.

\section*{Acknowledgment}
We thank Junlin Zhang for program management;
Joseph Heyward for helping with data processing;
Jean-Baptiste Alayrac, Matthew Grimes, Zihui Xue, Kaiming He and Leonidas Guibas for helpful feedback and discussions.

{
\small
\bibliographystyle{ieeenat_fullname}
\bibliography{main}
}
\clearpage
\clearpage
\maketitlesupplementary
\appendix

\startcontents[appendix]
\section*{Table of Contents}
\printcontents[appendix]{l}{1}{\setcounter{tocdepth}{2}}

\section{Dataset Details}
\label{appendix:dataset}

\subsection{Overview of single-life datasets.}
Table~\ref{tab:video_specs} details the total duration for each of the single life datasets used in our experiments.
Our single life training data is sourced from three main datasets: HD-Epic~\cite{perrett2025hdepic}, Walking Tours~\cite{venkataramanan2024dora}, and a private Anonymous Lives Dataset (ALD). We also include a control group of four ``non-life'' videos. For evaluating model alignment (CAS), we hold out one life from each of the main datasets, (WT9, P03, and Life5) for testing.

\begin{figure*}[p]
    \centering
    \begin{subfigure}[t]{0.9\linewidth}
        \centering
        \includegraphics[width=0.9\linewidth, height=0.21\textheight, keepaspectratio]{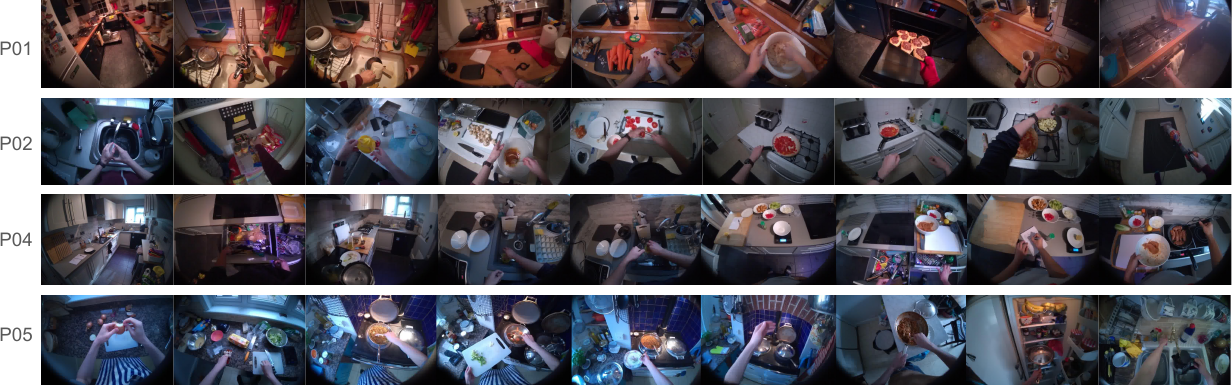}
        \caption{HD-Epic}
        \label{sup:frames_hdepic}
    \end{subfigure}
    \hfill
    \begin{subfigure}[t]{0.9\linewidth}
        \centering
        \includegraphics[width=0.9\linewidth, height=0.21\textheight, keepaspectratio]{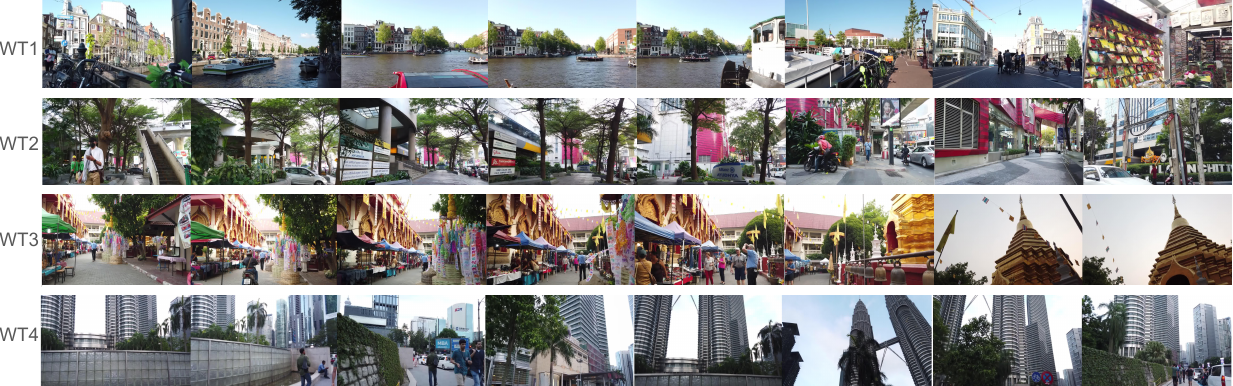}
        \caption{Walking Tours}
        \label{sup:frames_wt}
    \end{subfigure}

    \begin{subfigure}[t]{0.9\linewidth}
        \centering
        \includegraphics[width=0.81\textwidth, keepaspectratio]{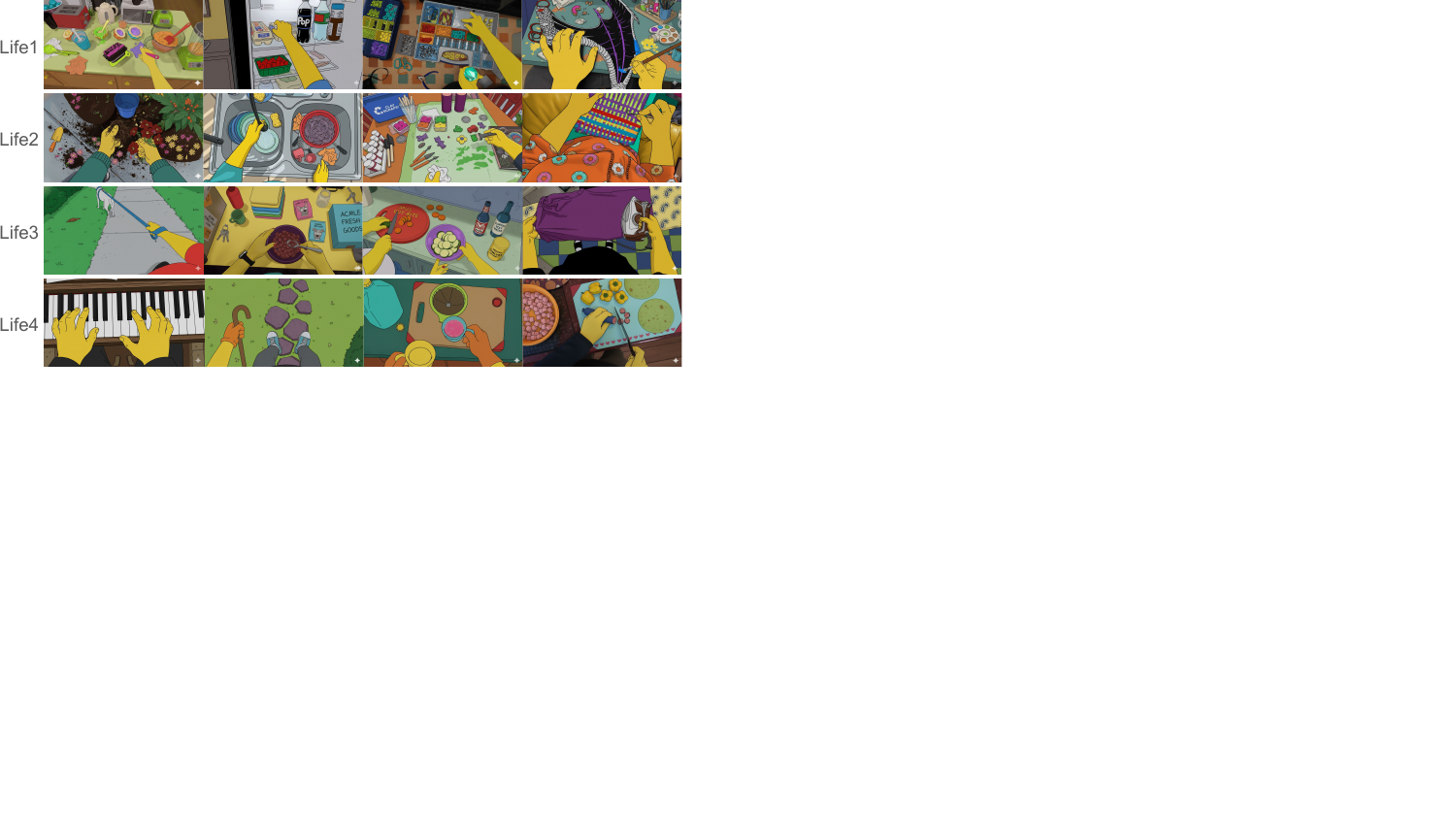}
        \caption{ALD (stylized for anonymity)}
        \label{sup:frames_ald}
    \end{subfigure}
    \hfill
    \begin{subfigure}[t]{0.9\linewidth}
        \centering
        \includegraphics[width=0.9\linewidth, height=0.21\textheight, keepaspectratio]{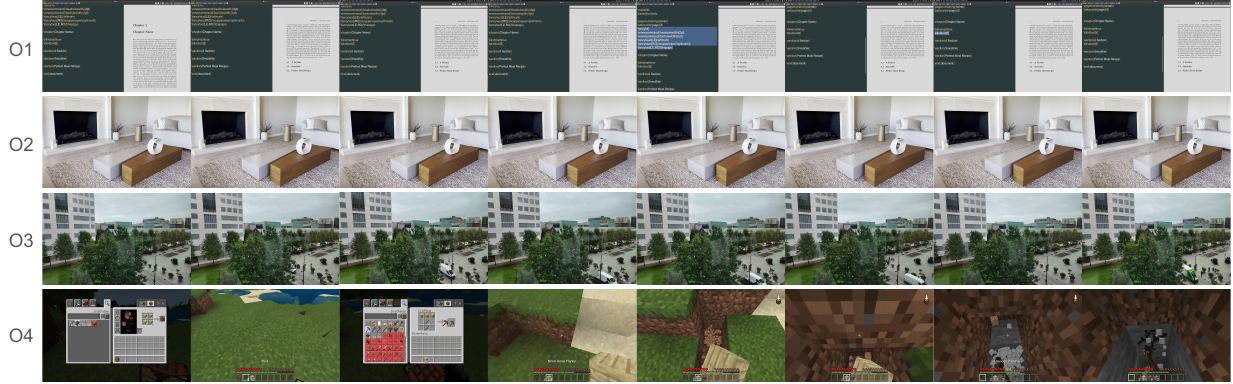}
        \caption{Other non-life videos}
        \label{sup:frames_other}
    \end{subfigure}
    \caption{A collection of sample frames for each of the datasets.
    }
    \label{fig:all_frames}
\end{figure*}

Figure~\ref{fig:all_frames} shows sample video frames from a few single lives. Due to privacy constraints, frames from private ALD datasets have been stylized for visualization using Gemini 2.5 Flash Image editing model (Nano-Banana). We emphasize that all models were trained exclusively on the original unstylized video frames. 

\begin{table}[b!]
\centering
\caption{Dataset duration for each life. `*' denotes testing lives.}
\label{tab:video_specs}
\resizebox{\columnwidth}{!}{%
\begin{tabular}{ll|ll|ll|ll}
\toprule
\textbf{HD-Epic} & \textbf{Dur.} & \textbf{WalkingTours} & \textbf{Dur.} & \textbf{ALD} & \textbf{Dur.} & \textbf{Other} & \textbf{Dur.} \\
\midrule
P01   & 5.09h & WT1: Amsterdam      & 1.36h &  Life1  &  30.5h & O1 & 1h \\
P02   & 4.58h & WT2: Bangkok        & 2.92h &  Life2  &  36.0h & O2 & 1h \\
P03* & 7.15h & WT3: Chiang Mai     & 1.13h &  Life3  &  30.4h & O3 & 1h \\
P04   & 4.62h & WT4: Kuala Lumpur   & 1.21h &  Life4  &  37.7h & O4 & 1h \\
P05   & 3.45h & WT5: Singapore      & 1.61h &  Life5* &  34.7h \\
P06   & 4.09h & WT6: Stockholm      & 1.11h &    &    \\
P07   & 3.59h & WT7: Venice         & 1.83h &    &    \\
P08   & 4.04h & WT8: Zurich         & 1.08h &    &    \\
P09   & 4.71h & WT9: Istanbul* & 1.13h &    &    \\
\bottomrule
\end{tabular}}
\end{table}

Table~\ref{tab:image_pairs_stats} summarizes the number of unique image pairs used for pre-training each of our single-life models. For all lives across all datasets, we report the number of pairs sampled using temporal pairing strategy. For the HD-Epic dataset, which contains the necessary ground truth geometry, we have also included the number of pairs sampled via spatial pairing and the union of both strategies. 

\begin{table}[t]
    \centering
    \scriptsize
    \caption{Number of sampled frame pairs for each life with different pairing strategies.}
    \label{tab:image_pairs_stats}
    \begin{tabular}{lccc}
        \toprule
        \textbf{Life Name} & \textbf{\#Temporal} & \textbf{\#Spatial} & \textbf{\#Union} \\
        \midrule
        P01 & 2,734,960 & 1,966,080 & 1,911,509 \\
        P02 & 2,165,619 & 978,195 & 953,120 \\
        P04 & 2,484,197 & 987,465 & 973,587 \\
        P05 & 1,848,841 & 718,144 & 703,284 \\
        P06 & 1,944,373 & 627,391 & 619,193 \\
        P07 & 1,933,987 & 1,261,987 & 1,228,048 \\
        P08 & 2,172,066 & 743,630 & 734,383 \\
        P09 & 2,532,718 & 407,913 & 868,444 \\
        \midrule
        WT1& 736,605 & N/A &N/A\\
        WT2& 1,574,775 & N/A &N/A\\
        WT3& 611,025 & N/A &N/A \\
        WT4& 655,200 & N/A &N/A \\
        WT5& 869,805 & N/A & N/A\\
        WT6& 598,200 & N/A & N/A\\
        WT7& 989,760 & N/A & N/A\\
        WT8& 584,670 & N/A & N/A\\
        \midrule
        Life1& 5,183,970 & N/A & N/A \\
        Life2& 5,494,905 & N/A & N/A \\
        Life3& 3,897,839 & N/A & N/A \\
        Life4& 5,076,709 & N/A & N/A \\        
        \midrule
        O1 & 537,285 & N/A &N/A \\
        O2 & 549,210 & N/A & N/A\\
        O3 & 539,865 & N/A & N/A\\
        O4 & 544,710 & N/A & N/A\\
        \bottomrule
    \end{tabular}
\end{table}

\begin{figure*}[ht!]
\centering
\begin{subfigure}[b]{0.18\textwidth}
    \centering
    \includegraphics[width=\linewidth]{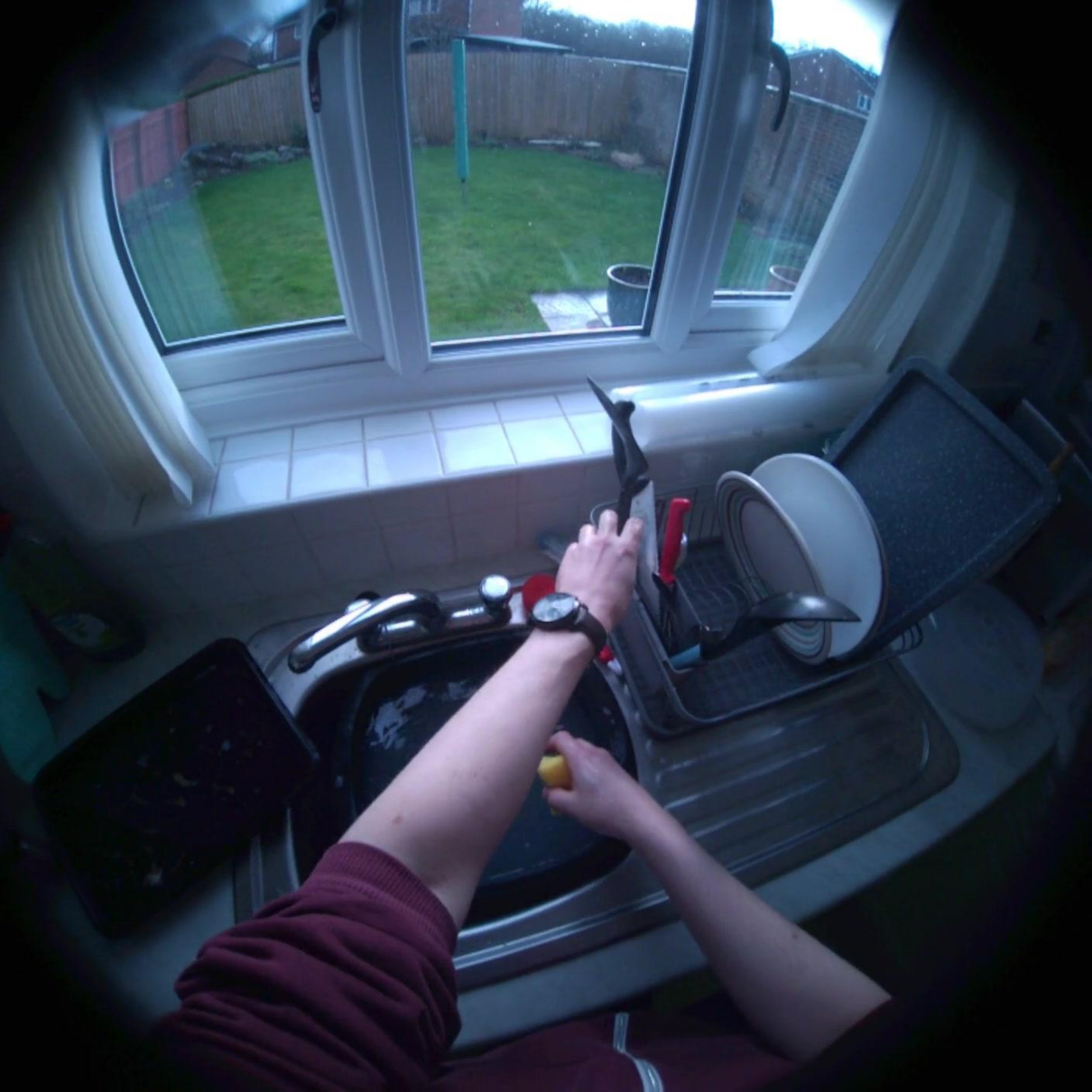}
    \caption{Source image ($I_t$)}
    \label{fig:pair_source}
\end{subfigure}%
\hfill
\begin{subfigure}[b]{0.18\textwidth}
    \centering
    \includegraphics[width=\linewidth]{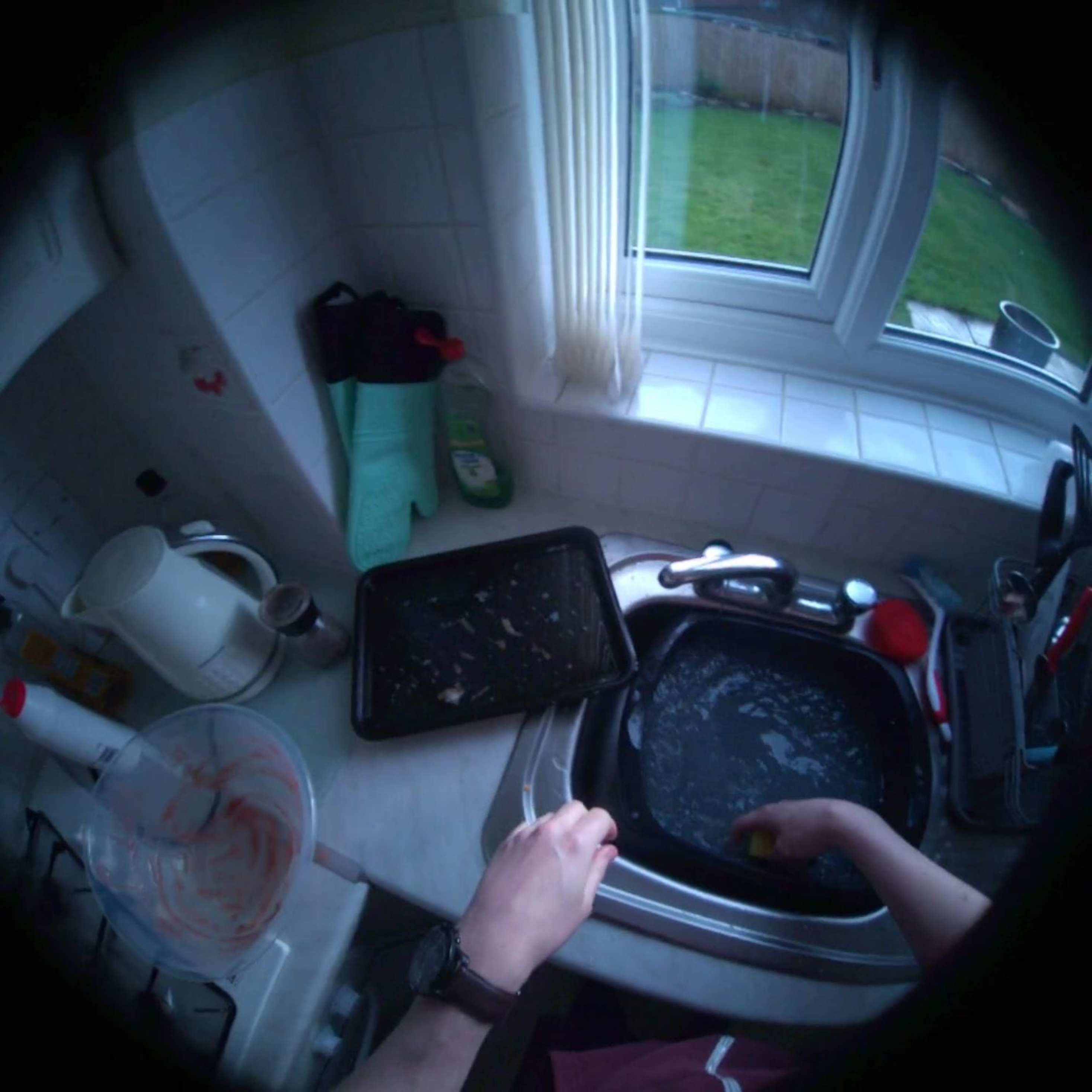}
    \caption{Temporal pair ($I_{t+\Delta t}$)}
    \label{fig:pair_temporal}
\end{subfigure}%
\hfill
\begin{subfigure}[b]{0.18\textwidth}
    \centering
    \includegraphics[width=\linewidth]{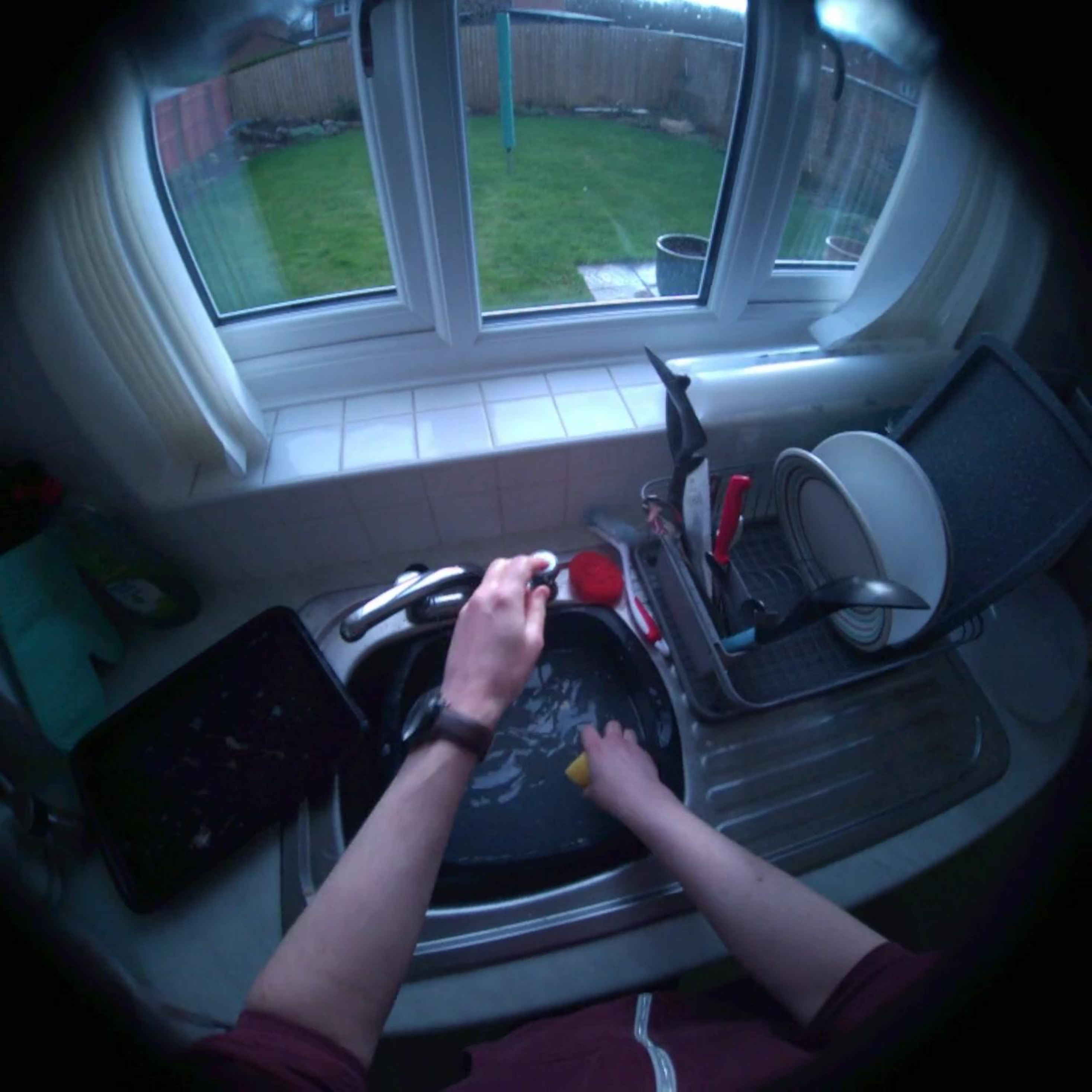}
    \caption{Spatial pair}
    \label{fig:pair_spatial}
\end{subfigure}%
\hfill 
\begin{subfigure}[b]{0.18\textwidth}
    \centering
    \includegraphics[width=\linewidth]{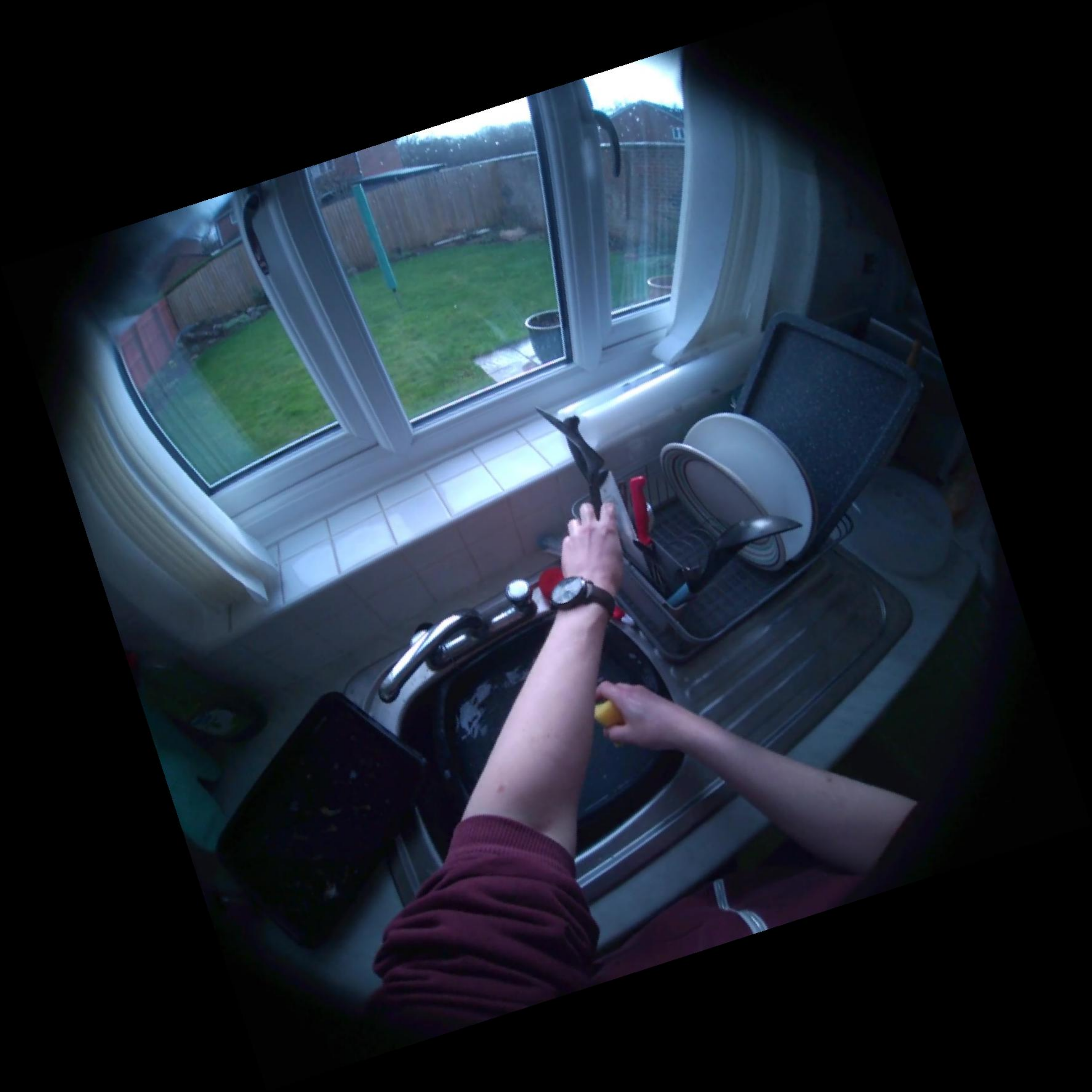}
    \caption{Augmented pair}
    \label{fig:pair_augmented}
\end{subfigure}%
\hfill
\begin{subfigure}[b]{0.18\textwidth}
    \centering
    \includegraphics[width=\linewidth]{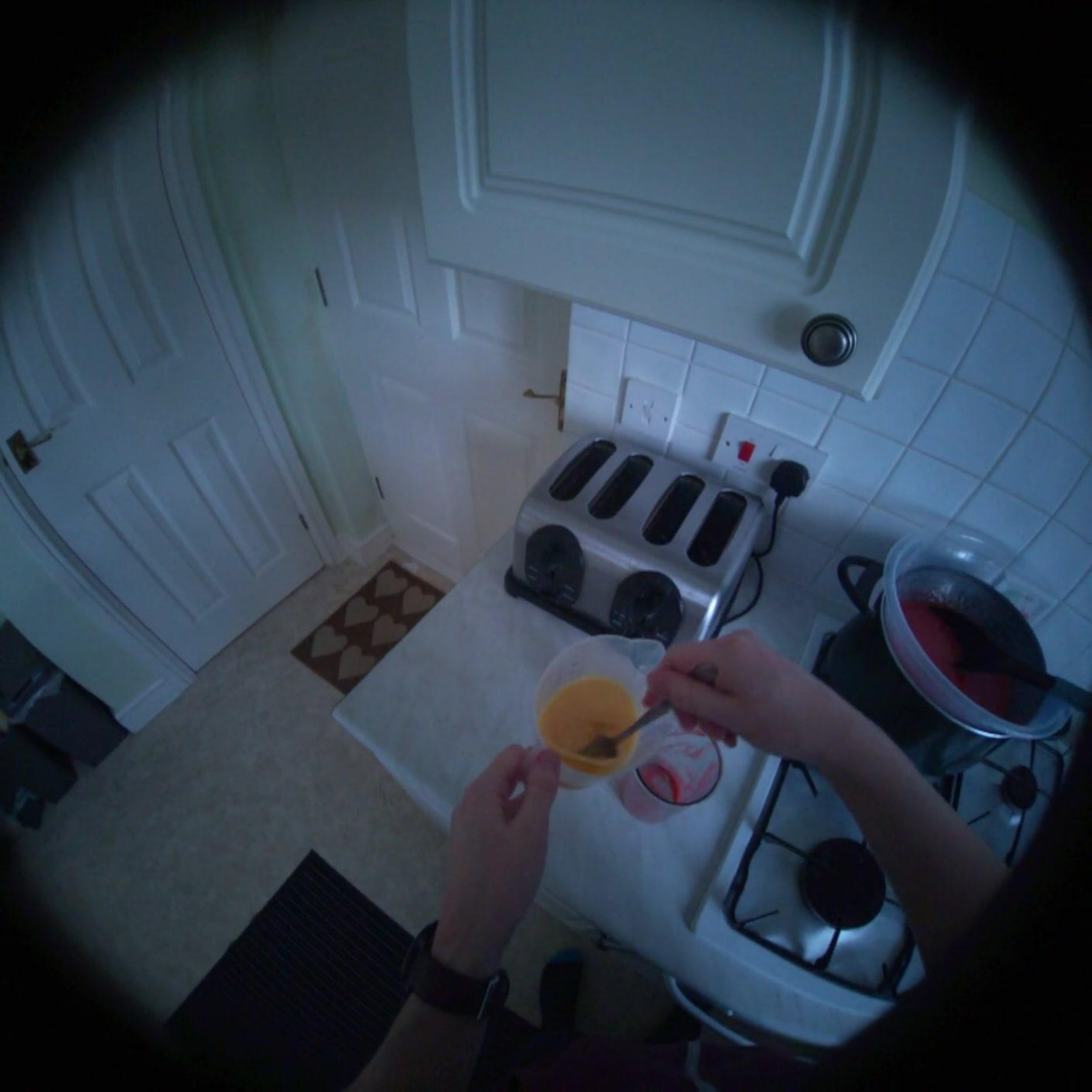}
    \caption{Random pair}
    \label{fig:pair_random}
\end{subfigure}%
\vspace{2mm}
\caption{
    \textbf{Visualization of pairing strategies.} Given a \textbf{(a)} source image, we generate pairs using different strategies. \textbf{(b)} A \textbf{temporal pair} is a future frame from the video, capturing natural motion. \textbf{(c)} A \textbf{spatial pair} is a non-consecutive frame with high geometric overlap, found using camera poses. \textbf{(d)} An \textbf{augmented pair} is a simple 2D transformation of the source image itself.  \textbf{(e)} A \textbf{random pair} is randomly selected from the same life data.
}
\label{fig:pairing_strategies_viz}
\end{figure*}

\subsection{How we obtain dataset properties}
\label{appendix:sec:dataset_stats}
In the main paper, Fig.~\ref{fig:properties_of_lives} visualizes a few properties for every life to highlight the similarities and differences across individuals' experiences. Here we explain how these properties are computed.
Each property is computed using 1000 uniformly sampled frames (for depth, brightness,~\etc) or consecutive frame pairs (for camera pose and optical flow) from each life. 
In every case, the statistic is averaged across the sampled frames.

\paragraph{Camera pose.} We estimated the relative camera motion between each pair of consecutive frames using DUSt3R~\cite{wang2024dust3r}. 
The pair of frames is 0.1 second apart, and the DUSt3R model infers the relative camera transformation between these two frames. From this transformation, we compute the rotation angle.

\paragraph{Depth.} To compute depth statistics, we used the Depth Anything V2 model~\cite{yang2024depthv2}, selecting specialized versions for indoor and outdoor scenes. For indoor-dominant datasets (HD-Epic, ALD, O1 and O2), we used the \texttt{\footnotesize Depth-Anything-V2-Metric-Indoor-Large-hf} model, which is fine-tuned on the Hypersim dataset. For outdoor datasets (WT, O3 and O4), we applied the \texttt{\footnotesize Depth-Anything-V2-Metric-Outdoor-Large-hf} model, fine-tuned on Virtual KITTI.
Each model processed the sampled frames to produce a high-resolution, per-pixel metric depth map for our analysis.

\paragraph{Optical flow magnitude.} 
We utilized RAFT~\cite{teed2020raft} for optical flow estimation, using the weights trained on the Sintel dataset. Similar to camera pose estimation, the optical flow estimation is performed on pairs of consecutive frames. 
For each pair, the model infers a dense optical flow field, which is a 2D vector at each pixel location representing its displacement to the subsequent frame. 
We then compute the average optical flow magnitude by calculating the Euclidean norm of optical flow for each pixel then averaging across all the pixels.

\paragraph{Brightness.}
We first converted each sampled frame to grayscale. The brightness for a single frame is then calculated as the mean pixel intensity of the resulting grayscale image. This value is subsequently normalized to a range from 0 to 1 by dividing by 255.

\paragraph{Objects.}
We investigated the presence of objects in every life using two approaches.
On HD-Epic where the object bounding box annotations are available with the dataset, we simply count the occurrences of these objects from the annotations.
For WT and ALD where there are no object annotations,
we prompt Gemini 2.5 Pro to recognize the visible objects in video frames and aggregate the object counts.
We collect statistics on 10 objects of interest: phone, table, animal, t-shirt, people, plant, car, food, knife and drawer.
For the spider plot on the right of Fig.~\ref{fig:properties_of_lives},
the object frequencies are normalized to 0-1, with respect to the count of `plant' which appears to be the most frequent object among the 10 chosen objects.

\subsection{Visualization of pairs sampled in each dataset.}
 Figure~\ref{fig:pairing_strategies_viz} provides a visual comparison of the different pairing strategies used in our experiments. Given a source image (a), we create training pairs that provide distinct supervisory signals. A \textit{temporal pair} (b) is a frame captured a short time after the source, containing natural motion and small viewpoint changes. A \textit{spatial pair} (c) is a temporally distant frame with significant viewpoint change but high geometric overlap, found using camera poses. These two strategies aim to capture the rich spatiotemporal structure of a single life. As baselines, we use an \textit{augmented pair} (d), which is a simple 2D transformation of the source image and lacks true 3D viewpoint change, and a \textit{random pair} (e), a completely unrelated frame from the same life that lacks any correspondence. 

\section{Implementation Details}
\label{appendix:implementation}

\subsection{Architecture and general pre-training details}
\label{appendix:implementation:training}
We re-implemented the CroCo~\citep{weinzaepfel2022croco} architecture in JAX~\citep{jax2018github} using a $\text{ViT-Base/16}$ encoder to encode images of $256\times256$ with a patch size of $16\times16$ pixels. For the decoder, we use a series of $12$ blocks with $768$ dimensions and $12$ attention heads. We pre-trained each single-life within HD-Epic, Walking Tours, and ALD for $100$, $30$, and $10$ epochs and a total batch size of $256$, $64$, and $256$ respectively. Overall, each model is trained for roughly 200-250K iterations. 
We use AdamW optimizer~\citep{loshchilov2017decoupled} with a cosine learning rate schedule with a base learning rate of $1.5 \times 10^{-4}$ and a linear warm-up in the first $10\%$ of total number of epochs for each dataset. We train each model using $16$ TPUs-v6. We used random crop, resizing, and color augmentation. For temporal pairing we used $95\%$ masking ratio in all our pretraining experiments. We performed a sweep on masking ratio for spatial-based pairing experiments detailed in~\Cref{appendix:sweep}.

\subsection{Downstream evaluation details}

\paragraph{Depth estimation tasks.}
We use a consistent set of hyperparameters to evaluate downstream depth estimation on both NYU-Depth-V2 and ScanNet. A lightweight readout module, consisting of a single Transformer block (1024 hidden dimension, 16 heads) followed by two linear layers, is attached to the frozen pre-trained encoder. This module is trained for 40k iterations with a batch size of 32 using the AdamW optimizer. We use a cosine learning rate schedule with a peak learning rate of $3\times10^{-4}$. Additional results using Attentive Finetuning and a DPT head are available in Appendix~\ref{appendix:subsection:readout}.

\paragraph{Zero-shot correspondence on HPatches implementation details.}
Following the protocol of ZeroCo~\cite{an2025cross}, we evaluate zero-shot correspondence on the HPatches dataset. For each image pair, we compute a final correspondence map, $\mathcal{A}_i$, by averaging the cross-attention maps from all decoder blocks of our pre-trained model. Performance is measured by computing the Average End-Point Error (AEPE) on the HPatches-240 variant, where correspondences are evaluated at a $240 \times 240$ resolution.

\subsection{Life-size experiment details}

In the main paper, Fig.~\ref{fig:scale_croco} and~\ref{fig:downstream_scale} compare single-life models trained on different ``life sizes".
The term ``life-size'' refers to a fixed duration of videos from an individual's experience.
For example, to train a model with a life-size of $T$, we randomly sample a video segment of duration $T$ from a single-life's video.
Specifically, we vary the life size across the following durations:
$\{ 3.6{\text{s}}, 36\text{s}, 6\text{m}, 12\text{m}, 30\text{m}, 45\text{m}, 1\text{h}, 2\text{h}, 3\text{h}, 30\text{h} \}$ subject to the maximum available duration of each life. Since different lives naturally have different total lengths (e.g., HD-Epic is $\sim$3.5h, Walking Tours is $\sim$1.1h, and ALD is $\sim$30.5h), 
the curves in Fig.~\ref{fig:scale_croco} and~\ref{fig:downstream_scale} 
stop at different points. 
For our Kinetics baseline, which is composed of a large number of short, 10-second videos, we create subsets of its training videos to match the total duration of each life-size experiment. 

For this scaling analysis, all 
training hyperparameters were held constant: 
the models shown in Fig.~\ref{fig:scale_croco} and~\ref{fig:downstream_scale} were each trained for 250k iterations with batch size 64, with other hyperparameters matching those in
~\Cref{appendix:implementation:training}.

\subsection{Masking ratio and Jaccard threshold}\label{appendix:sweep}
To determine the optimal hyperparameters for our spatial-based pairing strategies, we conducted a grid search for each life within the HD-Epic dataset. We swept over the masking ratio ($m \in \{0.5, 0.7, 0.9\}$) and the Jaccard co-visibility threshold ($j \in \{0.5, 0.7, 0.9\}$) for both the spatial and union pairing methods. The performance of each configuration was evaluated via attentive probing on the ScanNet and NYU-Depth-v2 depth estimation tasks. The full results of this sweep are presented in Table~\ref{tab:hyperparameter_sweep}. The best-performing combination of masking ratio and Jaccard threshold is highlighted in bold for each life. For all main experiments involving these pairing strategies, we report results using the optimal hyperparameters selected on a per-life basis from this sweep.

\begin{table*}[ht!]
\centering
\scriptsize
\caption{
Hyperparameter sweep for spatial and union pairing strategies on the HD-Epic dataset. We varied the masking ratio (\textit{m}) and the Jaccard threshold (\textit{j}). Performance is measured by attentive probing on ScanNet (AbsRel $\downarrow$) and NYU-Depth-v2 ($\delta_1$ $\uparrow$). The best performing setting for each participant and strategy is highlighted in \textbf{bold}.
}
\label{tab:hyperparameter_sweep}
\sisetup{
  table-format=1.4,
  round-mode=places,
  round-precision=4
}
\begin{tabular}{ll l S S S S S S}
\toprule
\multirow{2}{*}{\textbf{PID}} & \multirow{2}{*}{\textbf{Pairing}} & \multirow{2}{*}{\textbf{Masking Ratio (\textit{m})}} & \multicolumn{2}{c}{\textbf{Jaccard (\textit{j}) = 0.5}} & \multicolumn{2}{c}{\textbf{Jaccard (\textit{j}) = 0.7}} & \multicolumn{2}{c}{\textbf{Jaccard (\textit{j}) = 0.9}} \\
\cmidrule(lr){4-5} \cmidrule(lr){6-7} \cmidrule(lr){8-9}
& & & {ScanNet $\downarrow$} & {NYUv2 $\uparrow$} & {ScanNet $\downarrow$} & {NYUv2 $\uparrow$} & {ScanNet $\downarrow$} & {NYUv2 $\uparrow$} \\
\midrule

\multirow{6}{*}{P01} & \multirow{3}{*}{Spatial} 
 & 0.5 & 0.23298 & 0.5694 & \textbf{0.22561} & \textbf{0.5766} & 0.22991 & 0.5868 \\
&& 0.7 & 0.24109 & 0.5694 & 0.23039 & 0.5795 & 0.23498 & 0.5773 \\
&& 0.9 & 0.23608 & 0.5772 & 0.23291 & 0.5736 & 0.24171 & 0.5675 \\
\cmidrule(lr){2-9}
& \multirow{3}{*}{Union}   
 & 0.5 & 0.23263 & 0.5737 & \textbf{0.22501} & \textbf{0.5818} & 0.23007 & 0.5759 \\
&& 0.7 & 0.23920 & 0.5701 & 0.22983 & 0.5791 & 0.23394 & 0.5710 \\
&& 0.9 & 0.23704 & 0.5703 & 0.23694 & 0.5731 & 0.24265 & 0.5684 \\
\midrule

\multirow{6}{*}{P02} & \multirow{3}{*}{Spatial} 
 & 0.5 & 0.23893 & 0.5708 & \textbf{0.24547} & \textbf{0.5738} & 0.25027 & 0.5760 \\
&& 0.7 & 0.24682 & 0.5651 & 0.24084 & 0.5683 & 0.25588 & 0.5685 \\
&& 0.9 & 0.24921 & 0.5607 & 0.24583 & 0.5634 & 0.27046 & 0.5418 \\
\cmidrule(lr){2-9}
& \multirow{3}{*}{Union}   
 & 0.5 & 0.22924 & 0.5681 & \textbf{0.22561} & \textbf{0.5773} & 0.23194 & 0.5746 \\
&& 0.7 & 0.23101 & 0.5691 & 0.22755 & 0.5787 & 0.24614 & 0.5539 \\
&& 0.9 & 0.23659 & 0.5579 & 0.24256 & 0.5593 & 0.25528 & 0.5391 \\
\midrule

\multirow{6}{*}{P04} & \multirow{3}{*}{Spatial} 
 & 0.5 & 0.31445 & 0.5073 & 0.25313 & 0.5676 & \textbf{0.24861} & \textbf{0.5677} \\
&& 0.7 & 0.30204 & 0.5093 & 0.26628 & 0.5650 & 0.27892 & 0.5607 \\
&& 0.9 & 0.26633 & 0.5495 & 0.26005 & 0.5541 & 0.25773 & 0.5481 \\
\cmidrule(lr){2-9}
& \multirow{3}{*}{Union}   
 & 0.5 & 0.26735 & 0.5146 & 0.23646 & 0.5617 & \textbf{0.23137} & \textbf{0.5666} \\
&& 0.7 & 0.26602 & 0.5286 & 0.24015 & 0.5580 & 0.24131 & 0.5652 \\
&& 0.9 & 0.24927 & 0.5497 & 0.24143 & 0.5545 & 0.24315 & 0.5559 \\
\midrule

\multirow{6}{*}{P05} & \multirow{3}{*}{Spatial} 
 & 0.5 & 0.24805 & 0.5618 & 0.25664 & 0.5547 & 0.25369 & 0.5537 \\
&& 0.7 & \textbf{0.25423} & \textbf{0.5733} & 0.25637 & 0.5641 & 0.25608 & 0.5635 \\
&& 0.9 & 0.26484 & 0.5619 & 0.27185 & 0.5520 & 0.27598 & 0.5390 \\
\cmidrule(lr){2-9}
& \multirow{3}{*}{Union}   
 & 0.5 & 0.23573 & 0.5590 & \textbf{0.23373} & \textbf{0.5692} & 0.24314 & 0.5592 \\
&& 0.7 & 0.24091 & 0.5600 & 0.23487 & 0.5658 & 0.25035 & 0.5495 \\
&& 0.9 & 0.24161 & 0.5635 & 0.24532 & 0.5590 & 0.26098 & 0.5400 \\
\midrule

\multirow{6}{*}{P06} & \multirow{3}{*}{Spatial} 
 & 0.5 & 0.26266 & 0.5449 & 0.26766 & 0.5416 & 0.26010 & 0.5572 \\
&& 0.7 & \textbf{0.25490} & \textbf{0.5711} & 0.25553 & 0.5628 & 0.26939 & 0.5585 \\
&& 0.9 & 0.26070 & 0.5651 & 0.27273 & 0.5542 & 0.27846 & 0.5330 \\
\cmidrule(lr){2-9}
& \multirow{3}{*}{Union}   
 & 0.5 & 0.25208 & 0.5467 & \textbf{0.23815} & \textbf{0.5700} & 0.24129 & 0.5629 \\
&& 0.7 & 0.25063 & 0.5485 & 0.23889 & 0.5616 & 0.24763 & 0.5489 \\
&& 0.9 & 0.24270 & 0.5600 & 0.24656 & 0.5561 & 0.26034 & 0.5351 \\
\midrule

\multirow{6}{*}{P07} & \multirow{3}{*}{Spatial} 
 & 0.5 & 0.25310 & 0.5564 & 0.25554 & 0.5523 & 0.25423 & 0.5580 \\
&& 0.7 & 0.24368 & 0.5675 & \textbf{0.24997} & \textbf{0.5666} & 0.25399 & 0.5590 \\
&& 0.9 & 0.24435 & 0.5646 & 0.24632 & 0.5668 & 0.26567 & 0.5552 \\
\cmidrule(lr){2-9}
& \multirow{3}{*}{Union}   
 & 0.5 & 0.23771 & 0.5486 & \textbf{0.22904} & \textbf{0.5730} & 0.23323 & 0.5652 \\
&& 0.7 & 0.24234 & 0.5469 & 0.23219 & 0.5685 & 0.23609 & 0.5681 \\
&& 0.9 & 0.23985 & 0.5676 & 0.23906 & 0.5641 & 0.24273 & 0.5603 \\
\midrule

\multirow{6}{*}{P08} & \multirow{3}{*}{Spatial} 
 & 0.5 & 0.33090 & 0.4601 & 0.32803 & 0.4608 & 0.28933 & 0.5055 \\
&& 0.7 & 0.24695 & 0.5559 & 0.25797 & 0.5403 & 0.25737 & 0.5409 \\
&& 0.9 & \textbf{0.24209} & \textbf{0.5558} & 0.24792 & 0.5475 & 0.25065 & 0.5515 \\
\cmidrule(lr){2-9}
& \multirow{3}{*}{Union}   
 & 0.5 & 0.33340 & 0.4610 & 0.24437 & 0.5489 & \textbf{0.24133} & \textbf{0.5538} \\
&& 0.7 & 0.32396 & 0.4715 & 0.25165 & 0.5415 & 0.24657 & 0.5470 \\
&& 0.9 & 0.29374 & 0.4999 & 0.25488 & 0.5426 & 0.25015 & 0.5501 \\
\midrule

\multirow{6}{*}{P09} & \multirow{3}{*}{Spatial} 
 & 0.5 & 0.26253 & 0.5496 & 0.25512 & 0.5508 & 0.25642 & 0.5586 \\
&& 0.7 & 0.24207 & 0.5707 & 0.24578 & 0.5691 & 0.25229 & 0.5514 \\
&& 0.9 & \textbf{0.25262} & \textbf{0.5718} & 0.25435 & 0.5617 & 0.26509 & 0.5463 \\
\cmidrule(lr){2-9}
& \multirow{3}{*}{Union}   
 & 0.5 & 0.23775 & 0.5501 & \textbf{0.22717} & \textbf{0.5793} & 0.23365 & 0.5741 \\
&& 0.7 & 0.24051 & 0.5583 & 0.23341 & 0.5657 & 0.24243 & 0.5631 \\
&& 0.9 & 0.23687 & 0.5620 & 0.24277 & 0.5577 & 0.25214 & 0.5451 \\
\bottomrule
\end{tabular}
\end{table*}

\subsection{CAS implementation details}
For CAS computation, we use $k=5$ for Mutual Top-$k$. A pseudo-code to compute CAS for one pair of images in NumPy style is shown in Algorithm~\ref{alg:cas_score}.

\begin{algorithm}[t]
\caption{CAS Score (NumPy Implementation)}
\label{alg:cas_score}
\begin{lstlisting}[language=Python]
import numpy as np

def get_cas_score(a_i, a_j, k=5):
    """
    Calculate Correspondence Agreement Score.
    a_i: Patch correspondence map of model i. [HW, HW]
    a_j: Patch correspondence map of model j. [HW, HW]
    k: The top-k value
    """
    N = a_i.shape[0] 
    
    # Initialize boolean masks
    pred_mask = np.zeros((N, N), dtype=bool)
    target_mask = np.zeros((N, N), dtype=bool)
    
    # Create row indices for broadcasting
    row_indices = np.arange(N)[:, None]
    
    # Populate masks
    pred_mask[row_indices, a_i] = True
    target_mask[row_indices, a_j] = True
    
    # Calculate intersection over k
    mtopk = (pred_mask & target_mask).sum(axis=1) / k
    cas = mtopk.mean()
    
    return cas
\end{lstlisting}
\end{algorithm}

\section{Additional Results}
\label{appendix:results}

\subsection{Varying test set temporal intervals} 
In Table~\ref{tab:intervals}, we evaluated generalization beyond the training distribution ($[1, 16]$ frames) by sampling $1000$ pairs per dataset at gaps of $[17, 32]$ and $[33, 64]$ frames. As expected, average CAS on these $3000$-pair sets decreases as gaps increase and overlap narrows. However, models maintain consistent relative alignment, validating the robustness of our metric.

\begin{table}[ht!]
    \centering
    \caption{Varying test set temporal intervals on all datasets.}
    \label{tab:intervals}
    \resizebox{\linewidth}{!}{
    \begin{tabular}{lcccc}
        \toprule
\textbf{Dataset} & \textbf{[1, 16]} & \textbf{[17, 32]} & \textbf{[33, 64]} \\ \midrule
HD-Epic & 0.483 $\pm$ 0.021 & 0.340 $\pm$ 0.015 & 0.277 $\pm$ 0.014 \\
WalkingTours & 0.452 $\pm$ 0.015 & 0.315 $\pm$ 0.011 & 0.264 $\pm$ 0.007 \\
ALD & 0.492 $\pm$ 0.053 & 0.353 $\pm$ 0.035 & 0.292 $\pm$ 0.026 \\
\bottomrule
\end{tabular}%
}
\end{table}

\subsection{Zero-shot correspondence on HPatches}
\begin{figure*}[ht!]
\centering

\begin{subfigure}[b]{0.32\textwidth}
    \centering
    \includegraphics[width=\linewidth]{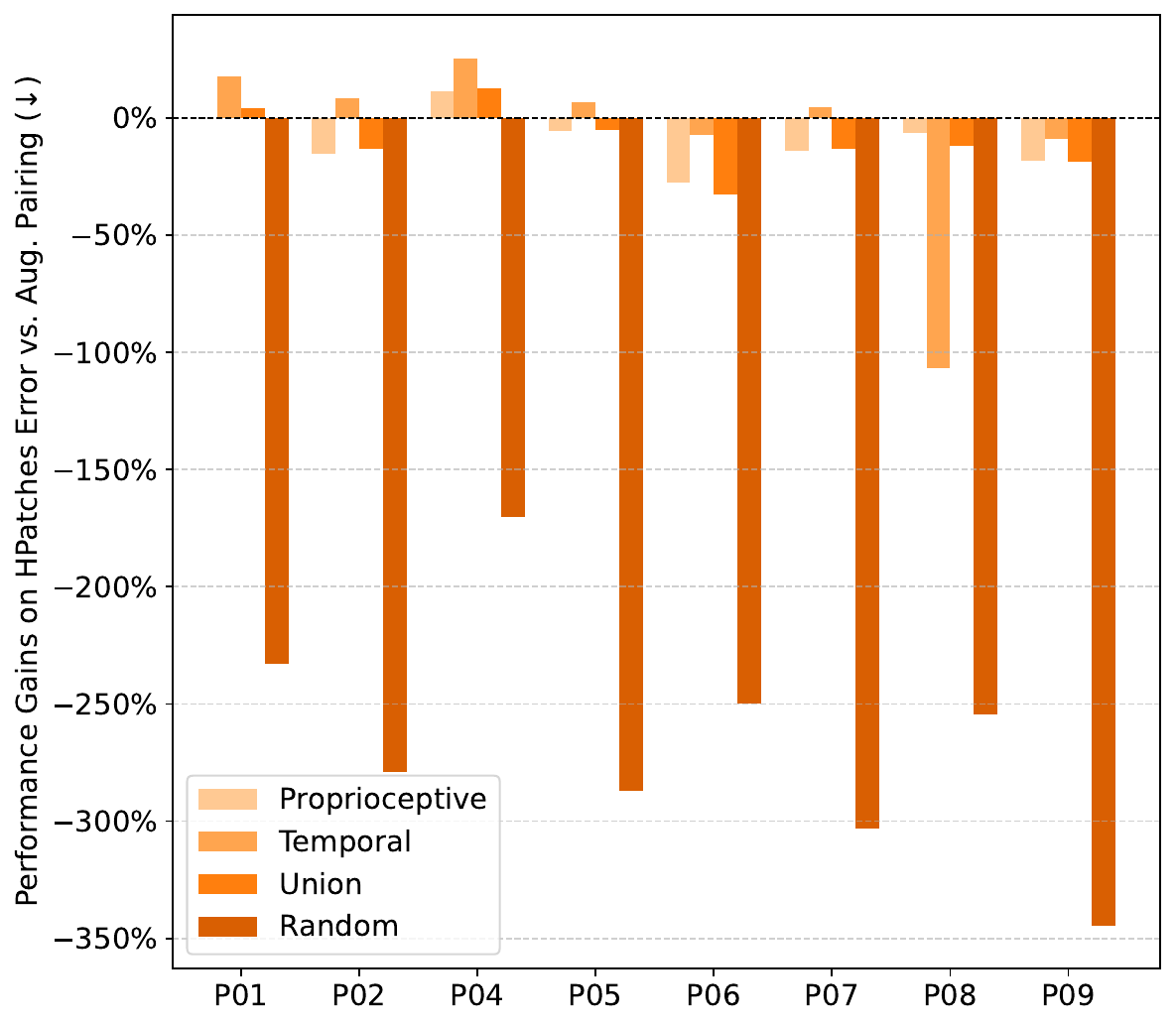}
    \caption{HD-Epic}
    \label{fig:hpatches_hdepic}
\end{subfigure}
\hfill
\begin{subfigure}[b]{0.32\textwidth}
    \centering
    \includegraphics[width=\linewidth]{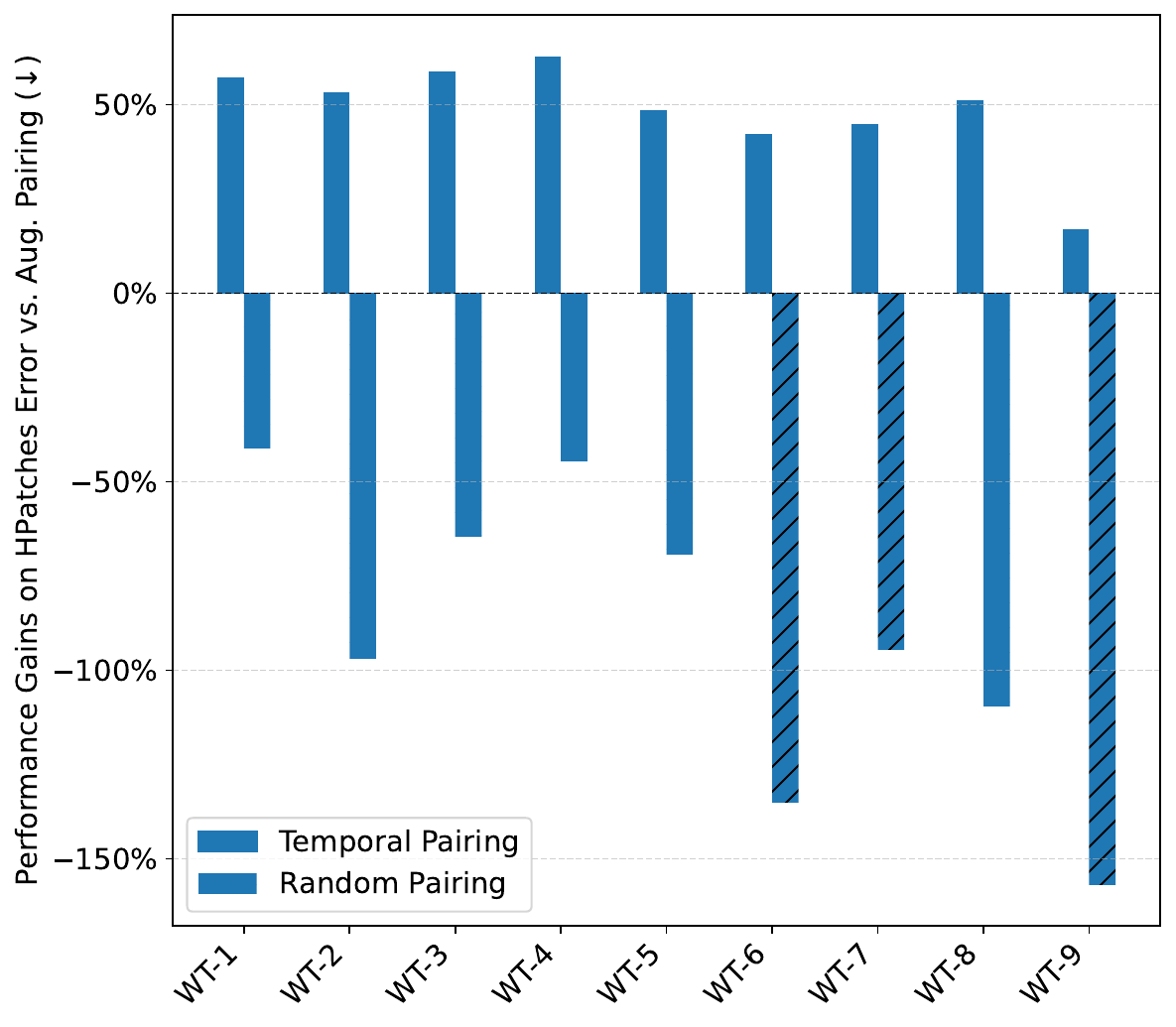}
    \caption{Walking Tours}
    \label{fig:hpatches_wt}
\end{subfigure}
\hfill
\begin{subfigure}[b]{0.32\textwidth}
    \centering
    \includegraphics[width=\linewidth]{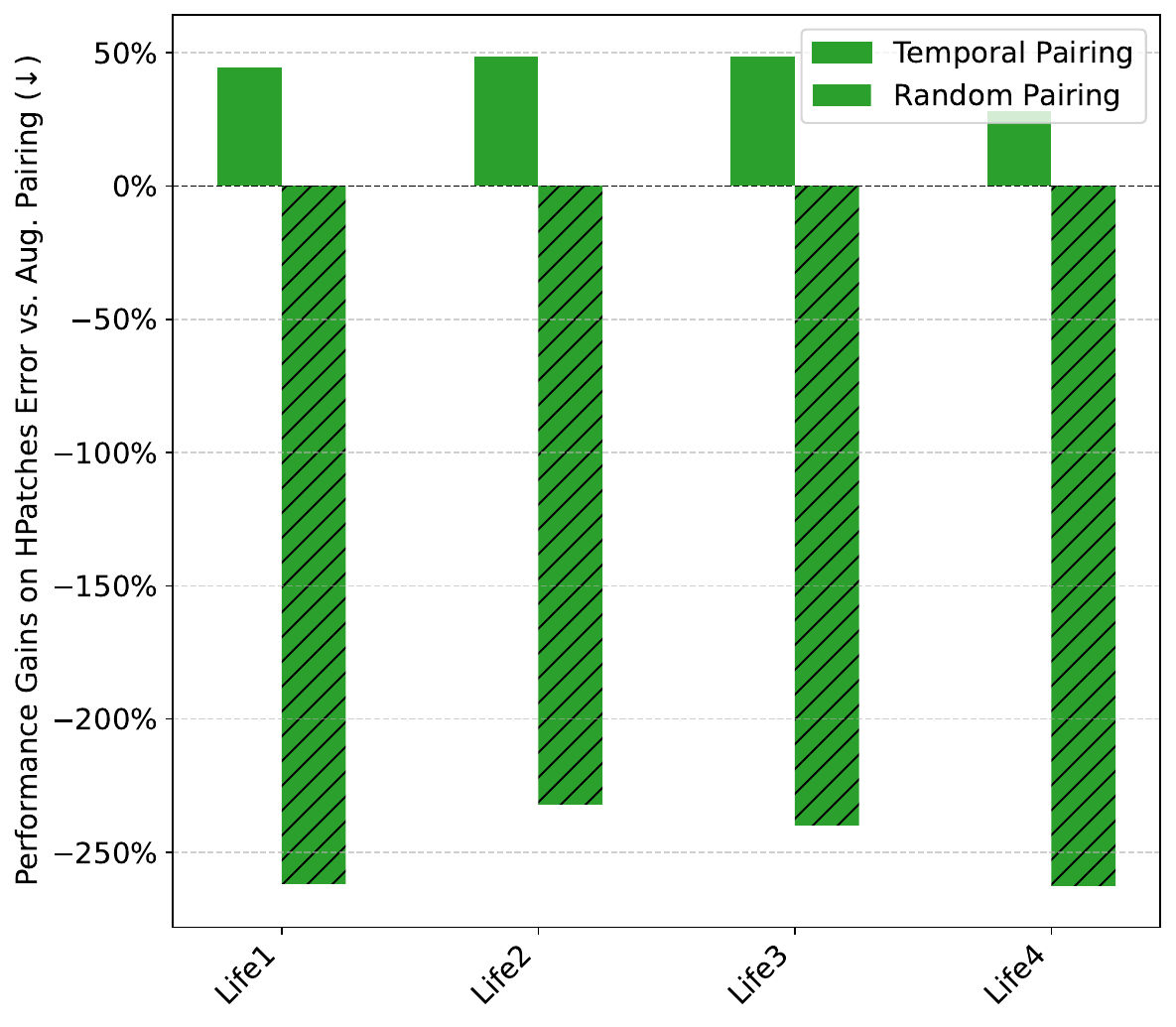}
    \caption{Anonymous Lives (ALD)}
    \label{fig:hpatches_ald}
\end{subfigure}
\caption{
    \textbf{Relative performance gains on zero-shot HPatches correspondence.} The plots show the percentage gain in performance (AEPE, lower is better) for different pairing strategies relative to the Augmented Pairing baseline (the 0\% line) for our three main datasets: \textbf{(a)} HD-Epic, \textbf{(b)} Walking Tours, and \textbf{(c)} Anonymous Lives.
}
\label{fig:hpatches_all_gains}
\end{figure*}

Figure~\ref{fig:hpatches_all_gains} 
shows the relative performance gains on this benchmark for all three of our main datasets. The results reinforce our findings from the depth estimation tasks: the Temporal Pairing strategy consistently and significantly outperforms both the Augmented and Random Pairing baselines across all datasets.

\subsection{Qualitative results on HPatches}

\begin{figure*}
    \centering
    \includegraphics[width=0.95\linewidth]{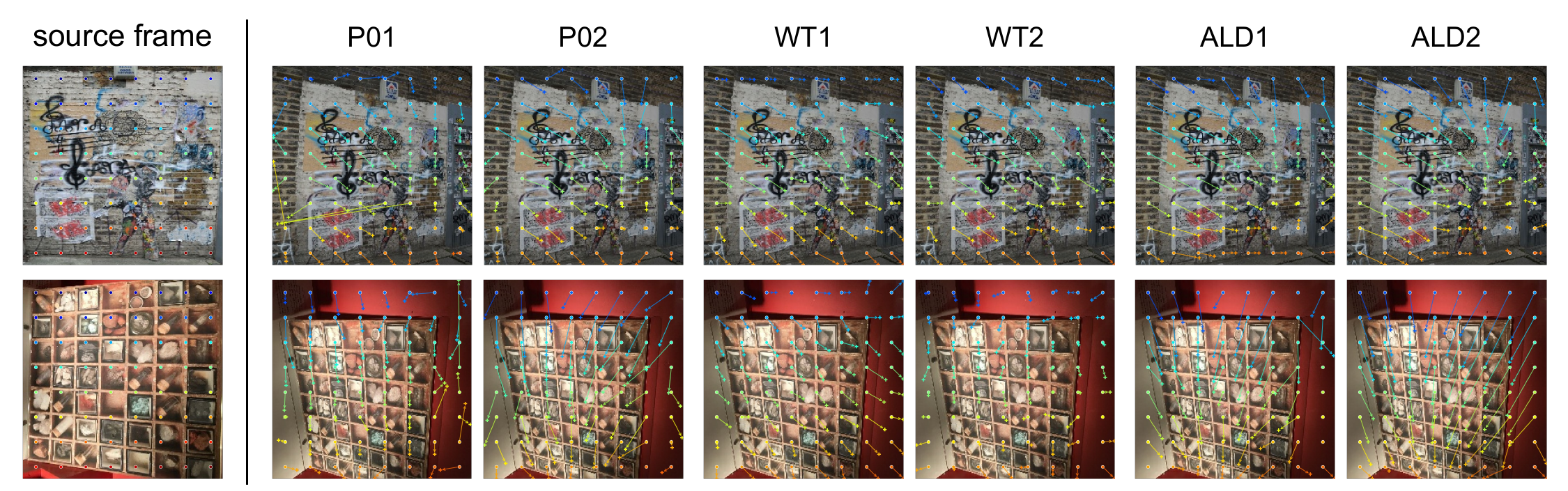}
    \caption{HPatches qualitative results from six single-life models. 
    We query a subset of source pixels (shown as ``dot''), and visualize the corresponding pixels in the target frames as ``$+$'' sign.
    The displacement of source pixels are visualized as arrows.}
    \label{fig:hpatch_qualitative}
\end{figure*}

Figure~\ref{fig:hpatch_qualitative} shows two HPatches samples for zero-shot correspondence task. Every single-life model takes both the source and target image as input, and for each pixel in the source frame, the model derives the corresponding pixels in the target frame. For clarity, we query $8\times8$ pixels in the source frame and visualize their corresponding pixels in the target frame.
The models trained on ALD lives perform better on this task, as the corresponding pixels in the target frames are more structured (the $+$ sign). Generally, all single-life models manage to find corresponding pixels to some extent.

\subsection{Other downstream evaluation protocols}
\label{appendix:subsection:readout}
On downstream depth estimation task,
we also experiment with different evaluation protocols other than Attentive probing.
For Attentive finetuning, we use the same single transformer block but finetune the whole network end-to-end for depth estimation task.
For DPT finetuning, we follow the original setup from CroCo~\cite{weinzaepfel2022croco} and use a DPT head~\cite{ranftl2021vision} attached on the pretrained encoder, and finetune the whole network end-to-end.

Table~\ref{tab:results} presents the comparison between Attentive Probing (Attn Frozen) and full Attentive Finetuning. As the results clearly indicate, allowing the entire network to finetune leads to substantial performance improvements across all datasets and tasks, with ScanNet evaluation on Walking Tours lives being the only case where the performance degrades. For example, on NYU-Depth-v2, finetuning improves $\delta_1$ accuracy by more than 10 percentage points on average, while on ScanNet, it reduces the absolute relative error by nearly 40\%. This significant performance boost confirms that the features learned from the single-life paradigm are not brittle; they serve as a strong and highly adaptable foundation for task-specific fine-tuning.

\begin{table}[t]
    \centering
    \caption{Different evaluation protocols on NYU-Depth-v2 depth estimation tasks. The $\delta_1$ accuracy is reported (higher the better).}
    \label{tab:probe_vs_ft}
\resizebox{0.6\linewidth}{!}{%
\begin{tabular}{lccc}
\toprule
\textbf{Dataset} & \textbf{Attn Frozen} & \textbf{Attn FT} & \textbf{DPT FT} \\
\midrule
{\color{gray}Habitat}~\cite{weinzaepfel2022croco} & {\color{gray}0.767} & {\color{gray}{0.828}} & {\color{gray}{0.892}} \\
K400 & 0.590 & 0.696 & 0.696 \\
\midrule
ALD-Life1 & 0.580 & 0.694 & 0.693 \\
ALD-Life2 & 0.580 & 0.684 & 0.705 \\
ALD-Life4 & 0.570 & 0.671 & 0.685 \\
\bottomrule
\end{tabular}%
}
\end{table}

\begin{table}[h]
    \centering
    \caption{Full finetuning results on ScanNet and NYUv2}
    \label{tab:results}
    \resizebox{\linewidth}{!}{
    \begin{tabular}{lcccc}
        \toprule
        & \multicolumn{2}{c}{\textbf{ScanNet} AbsRel($\downarrow$)} & \multicolumn{2}{c}{\textbf{NYUv2} $\delta_1$($\uparrow$)} \\
        \cmidrule(lr){2-3} \cmidrule(lr){4-5}
        \textbf{Dataset} & Attn Frozen & Attn Finetune & Attn Frozen & Attn Finetune \\
        \midrule
        HD-EPIC (Temporal) & $0.2569 \pm 0.0089$ & $0.1558 \pm 0.0096$ & $0.5617 \pm 0.0109$ & $0.6770 \pm 0.0148$ \\
        HD-EPIC (Spatial) & $0.2455 \pm 0.0096$ & $0.1545 \pm 0.0056$ & $0.5703 \pm 0.0063$ & $0.6867 \pm 0.0122$ \\
        WT (Temporal) & $0.2783 \pm 0.0079$ & $0.3451 \pm 0.0261$ & $0.5302 \pm 0.0124$ & $0.6410 \pm 0.0189$ \\
        ALD (Temporal) & $0.2377 \pm 0.0035$ & $0.1574 \pm 0.0039$ & $0.5795 \pm 0.0072$ & $0.6865 \pm 0.0094$ \\
        \bottomrule
    \end{tabular}
    }
\end{table}

\subsection{DINOv2 results and discussion}
Our experiments are conducted on the Cross-View Completion (CroCo) architecture.
To explore whether our findings extend to alternative self-supervised learning paradigms, we also evaluate on the DINOv2~\cite{oquab2023dinov2} training adapting it to our proposed `single-life' learning paradigm.
Specifically, we train several DINOv2 models (all using the same ViT-B architecture) \textit{from scratch} on individual lives, using the same single-life data as in the main paper.

There are some notable differences between DINOv2 and CroCo:
(1) DINOv2 takes single images as input. We extract video frames and shuffle them, thus creating an image dataset.
(2) DINOv2 consists of a single transformer encoder (without decoder). As a result, we adapt our score (CAS) to this scenario, and compute this score from the encoder feature correlation (used in ZeroCo~\cite{an2025cross}) rather than the decoder cross-attention maps. In other words, given a pair of images, we use patchwise encoder features to compute patch-to-patch similarity matrix $A_i$ associated with each trained model $i$. We then use these matrices to compute the CAS score using Eq. \eqref{alg:cas_score} defined in the main paper. This allows us to compare different trained DINOv2 models in the same way as we did for CroCo.

\begin{figure}[ht]
    \centering
        \includegraphics[width=0.98\linewidth]{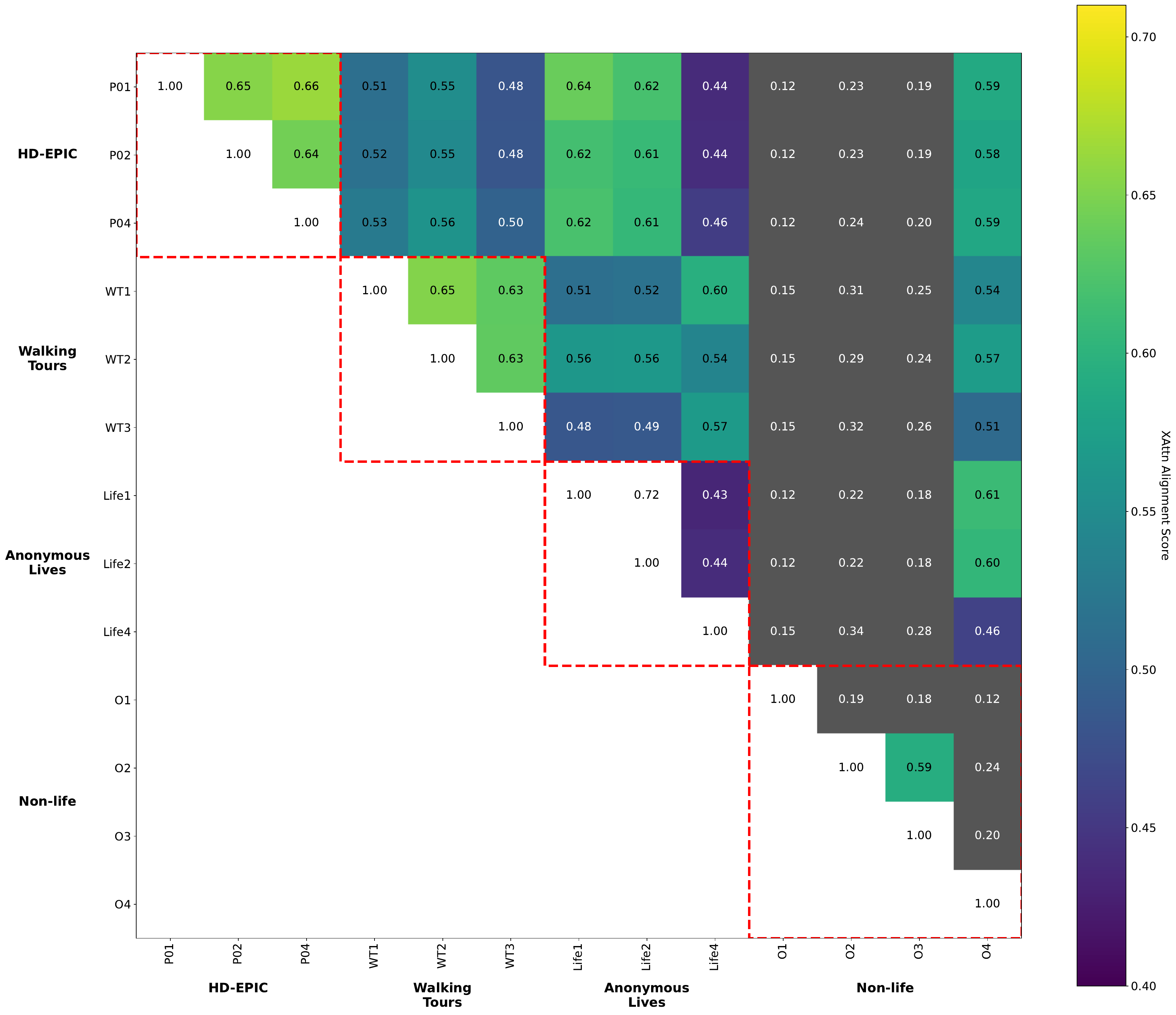}
    \caption{
    DINOv2 CAS metric across models. We do not observe a similar cross-model alignment within datasets like we observed for CroCo (red blocks). 
    }
    \label{fig:dino_matrix}
\end{figure}

\begin{figure}
    \centering
    \includegraphics[width=0.98\linewidth]{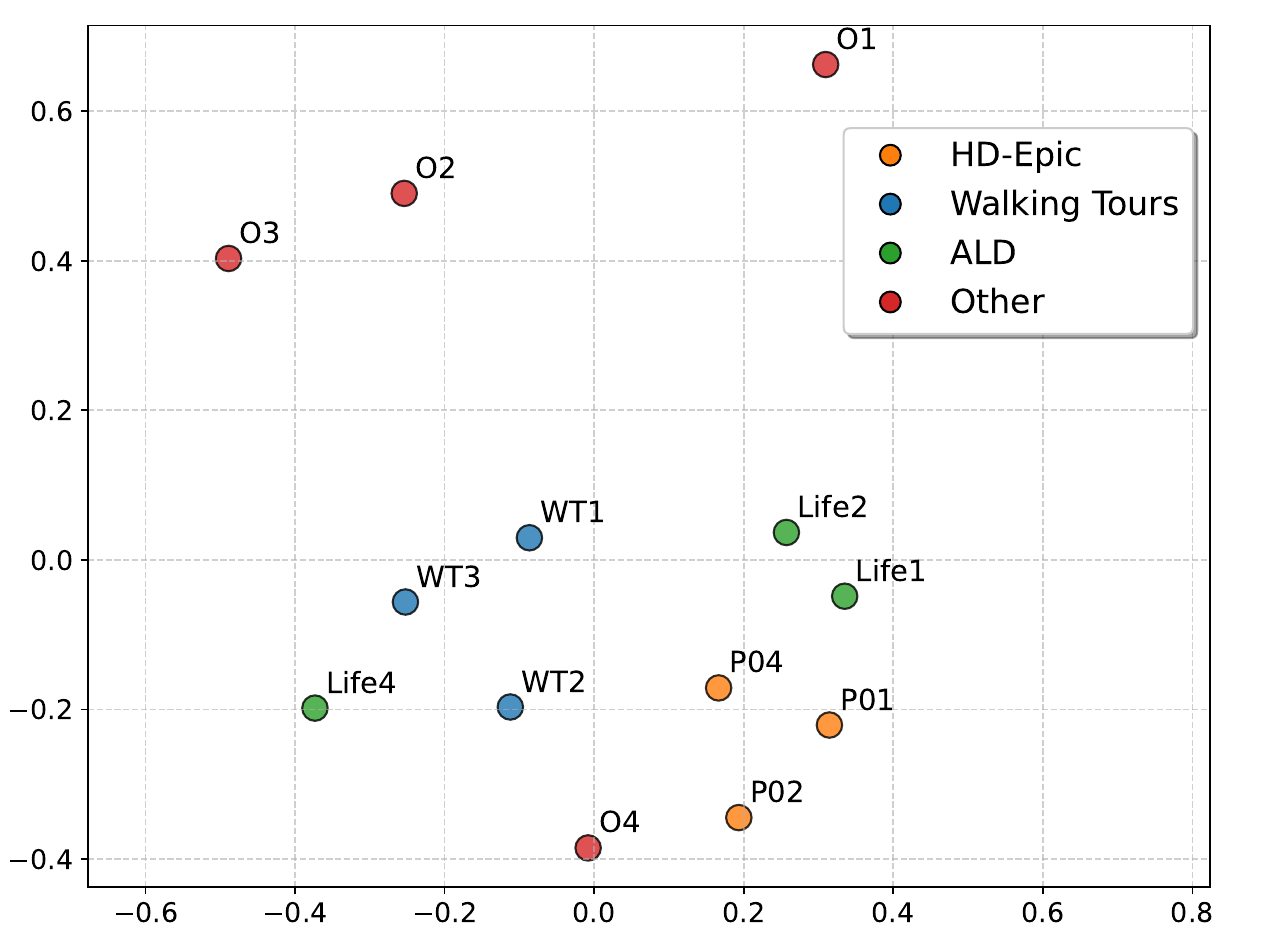}
    \caption{A 2D MDS visualization of the single-life DINOv2 models using the CAS score from Fig.~\ref{fig:dino_matrix} as the similarity metric.}
    \label{fig:mds_dino}
\end{figure}

Table~\ref{table:dinov2_results} summarizes the single-life DINOv2 results.
We find single-life DINOv2 gives similar depth estimation and zero-shot correspondence performance compared with single-life CroCo.
Notably, DINOv2 could not be trained stably on redundant lives (O2, O3) or out-of-distribution video (O1) - the training quickly gets infinite or NaN loss, 
indicating DINOv2 method is better suited for IID data and not robust to redundant input. We trained DINOv2 on a 30 hours long subset of K400 which achieves 63\% on NYU-Depth-v2, 0.193 on ScanNet, and 18.0 on HPatches further supporting DINOv2's suitability for IID-like data. We also present the cross-model CAS matrix for single-life DINOv2 models in Fig.~\ref{fig:dino_matrix}. Note that CAS is computed on the encoder feature correlation. Comparing with the main paper Fig.~\ref{fig:cas_matrix_a}, we observe that CAS metric results have higher variability for DINOv2: we do not observe a strong CAS intra-datasets compared to inter-datasets, and interestingly on O4 (minecraft) we find high similarity with other single-life models trained on natural lives. Note that O4 (minecraft) serves as a deliberate edge case, pairing first-person 3D consistency with non-realistic, pixelated visuals.
We also visualized a 2D MDS plot similar to the main paper Fig.~\ref{fig:cas_matrix_b} but for single-life DINOv2 models, in Fig.~\ref{fig:mds_dino}.

Our findings highlight that it is possible to train diverse architectures on single-life data. However, the combination of the single-life learning paradigm and a suitable training objective is important for learning representations that are both generalizable and consistently aligned. 

\begin{table}[h!]
    \centering
    \caption{Single-life DINOv2 results. 
    *DINOv2 training is unstable and does not converge on these datasets, hence we took the last checkpoint before Inf/NaN appears.}
    \label{table:dinov2_results}
    \resizebox{0.95\linewidth}{!}{%
    \begin{tabular}{l|cccc}
    \toprule
    \textbf{Dataset} & \textbf{NYUv2$\uparrow$}	& \textbf{ScanNet$\downarrow$} & \textbf{HPatches$\downarrow$} & \textbf{CAS w. CroCo$\uparrow$} \\ \midrule
    HD-Epic: $\{1,2,4\}$ & 0.589$\pm$0.009 & 0.222$\pm$0.005 & 19.8$\pm$1.5 & 0.536$\pm$0.01 \\
    WT: $\{2,3,7\}$      & $0.562\pm0.026$
    & 0.256$\pm$0.003 & $21.2\pm4.16$ 
    & 0.431$\pm$0.04 \\
    ALD: $\{1,2,4\}$     & 0.598$\pm$0.05  & 0.215$\pm$0.033 & 19.7$\pm$9.9    & 0.516$\pm$0.13     \\
    \midrule
    K400-30h & 0.632 & 0.193 & 18.0 & 0.524 \\
    \midrule
    O1* & 0.423 &  0.351 & 57.3  & 0.120 \\
    O2* & 0.453 &  0.326	& 51.1	& 0.210 \\
    O3* & 0.450 &  0.325	& 53.0	& 0.182 \\
    O4 & 0.596 &  0.217	& 20.4	& 0.514 \\
    \bottomrule
    \end{tabular}}
\end{table}

\subsection{Results beyond Geometric Tasks}
To assess whether our geometry-focused pre-training maintains semantic coherence, we qualitatively analyze performance on a \textbf{zero-shot segmentation label propagation task}~\cite{jabri2020walk} on \textbf{DAVIS 2017}~\cite{pont20172017}. Specifically, the model is given the ground-truth masks for multiple objects in the first frame and the task is to propagate these masks through the remaining frames of the video. Following common practice, all evaluations are performed on 480p resolution images. To isolate the quality of our learned representation, we employ a simple K-nearest neighbor inference algorithm. The features from our encoder are used to compute a dense similarity map between pixels, allowing labels to be propagated based on the $k$ most similar pixels in the source frame(s). This zero-shot approach directly probes the representation's utility for object-level correspondence without any task-specific fine-tuning.

\begin{figure*}[ht]
    \centering
        \includegraphics[width=0.9\linewidth]{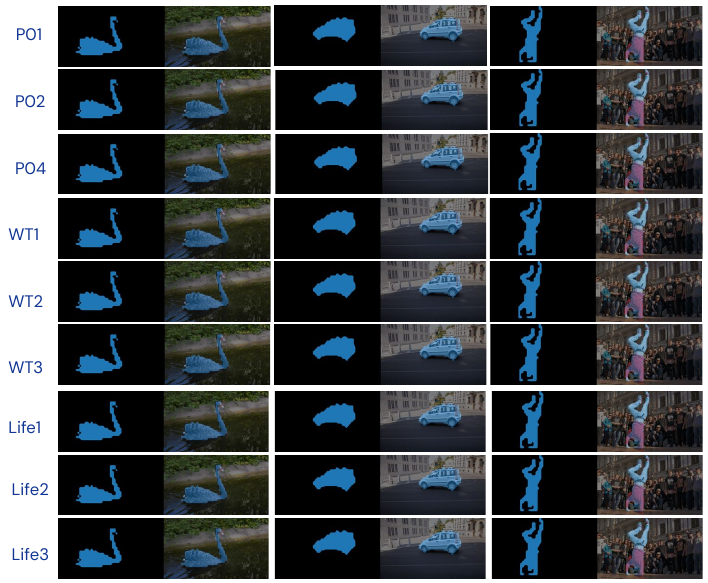}
    \caption{
    Video segmentation tracking results on DAVIS 2017 dataset from three single-life models. From top: HD-Epic P01,P02,P04, WT1,2,3, and ALD Life1,2,3. The features produced by the models are used to propagate the ground-truth segmentation labels of the first video frames (left) to the future frames (right) in a zero-shot manner.
    }
    \label{fig:segmentation}
\end{figure*}

As shown in Fig.~\ref{fig:segmentation}, models trained on three single lives from all three datasets show consistent and reasonable label propagation. 
These promising results highlight that these models, trained independently on different lives, can generalize to unseen objects -- e.g. none of our lives have seen a `swan' and all our indoor lives have not seen a `car' during training.

Additionally, we evaluate the models on the Perception Test point tracking task \cite{patraucean2023perception}, following the protocol in \cite{carreira2024scaling}. Using the HD-Epic single-life checkpoints, we observe that performance improves with increased data duration. Scaling the training data from 30 minutes to 1 hour and 3 hours improves the Average Jaccard (AJ) from 51\% to 54\% and 57\%, respectively. This upward trajectory approaches the 64\% AJ baseline achieved by a model trained on 850 hours of diverse K400 data, further validating the efficacy of the learning signal inherent in single-life videos.

\end{document}